\documentclass[10pt,twocolumn,letterpaper]{article}

\usepackage{cvpr}              

\usepackage{amsmath}
\usepackage{amsfonts}
\usepackage{multirow}
\usepackage{xcolor}

\newcommand{\methodname}{\text{NoPe-NeRF}\xspace}

\def\ie{\textit{i.e.}, }
\def\eg{\textit{e.g.}}

\usepackage{graphicx}
\usepackage{amsmath}
\usepackage{amssymb}
\usepackage{booktabs}
\usepackage{makecell}
\usepackage{tabularx}
\usepackage{rotating}
\usepackage[normalem]{ulem}
\usepackage[title]{appendix}
\usepackage[pagebackref,breaklinks,colorlinks]{hyperref}

\usepackage[capitalize]{cleveref}
\crefname{section}{Sec.}{Secs.}
\Crefname{section}{Section}{Sections}
\Crefname{table}{Table}{Tables}
\crefname{table}{Tab.}{Tabs.}
\usepackage[accsupp]{axessibility}
\def\cvprPaperID{2763} 
\def\confName{CVPR}
\def\confYear{2023}

\begin{document}

\title{NoPe-NeRF: Optimising Neural Radiance Field with No Pose Prior}

\author{Wenjing Bian \and Zirui Wang \and Kejie Li \and Jia-Wang Bian \and Victor Adrian Prisacariu\\
Active Vision Lab, University of Oxford\\
{\tt\small \{wenjing, ryan, kejie, jiawang, victor\}@robots.ox.ac.uk}
}
\twocolumn[{%
\renewcommand\twocolumn[1][]{#1}%
\maketitle
\begin{center}
    \centering
\includegraphics[width=0.96\linewidth]{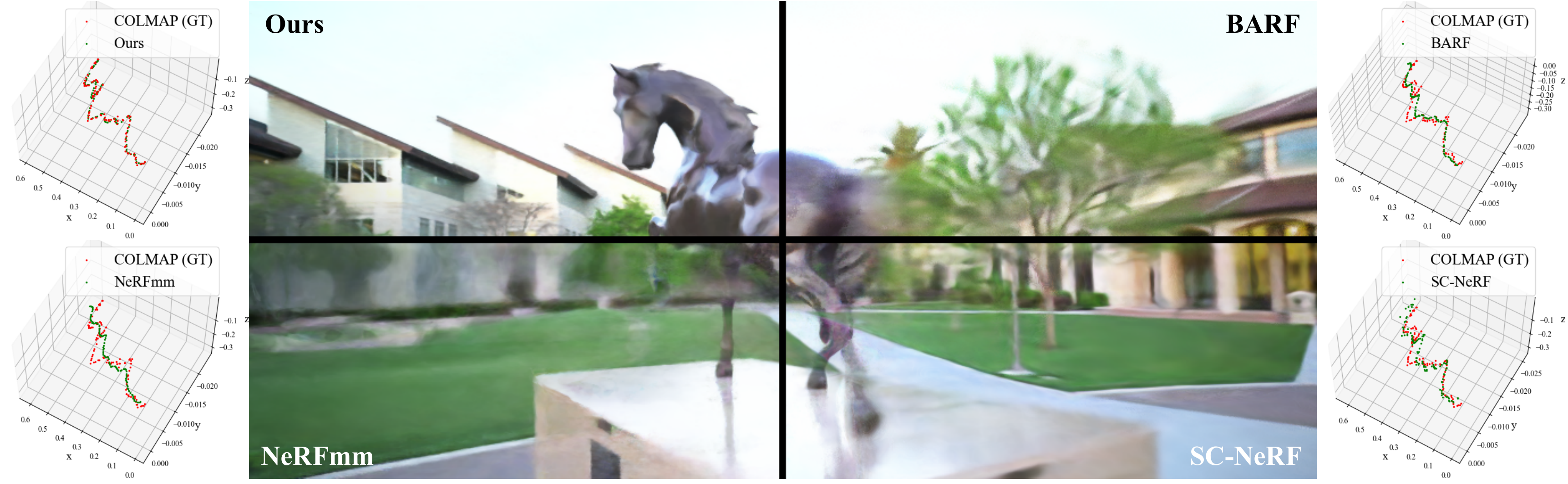}
    \captionof{figure}{\textbf{Novel view synthesis comparison}. 
    We propose \methodname for joint pose estimation and novel view synthesis. 
    Our method enables more robust pose estimation and renders better novel view synthesis than previous state-of-the-art methods. 
    }
\end{center}\label{fig:page1}
}]


\begin{abstract}
Training a Neural Radiance Field (NeRF) without pre-computed camera poses is challenging. Recent advances in this direction demonstrate the possibility of jointly optimising a NeRF and camera poses in forward-facing scenes. However, these methods still face difficulties during dramatic camera movement. We tackle this challenging problem by incorporating undistorted monocular depth priors. These priors are generated by correcting scale and shift parameters during training, with which we are then able to constrain the relative poses between consecutive frames. This constraint is achieved using our proposed novel loss functions. Experiments on real-world indoor and outdoor scenes show that our method can handle challenging camera trajectories and outperforms existing methods in terms of novel view rendering quality and pose estimation accuracy. Our project page is \url{https://nope-nerf.active.vision}.
\end{abstract}
\section{Introduction} \label{sec:intro}
The photo-realistic reconstruction of a scene from a stream of RGB images requires both accurate 3D geometry reconstruction and view-dependent appearance modelling.
Recently, Neural Radiance Fields (NeRF)~\cite{mildenhall2021nerf} have demonstrated the ability to build high-quality results for generating photo-realistic images from novel viewpoints given a sparse set of images. 


An important preparation step for NeRF training is the estimation of camera parameters for the input images. A current go-to option is the popular Structure-from-Motion (SfM) library COLMAP~\cite{schonberger2016structure}.
Whilst easy to use, this preprocessing step could be an obstacle to NeRF research and real-world deployments in the long term due to its long processing time and its lack of differentiability. Recent works such as NeRFmm~\cite{wang2021nerfmm}, BARF~\cite{lin2021barf} and SC-NeRF~\cite{jeong2021self} propose to simultaneously optimise camera poses and the neural implicit representation to address these issues.
Nevertheless, these methods can only handle forward-facing scenes when no initial parameters are supplied, and fail in dramatic camera motions, \eg  a casual handheld captured video.

This limitation has two key causes. 
First, all these methods estimate a camera pose for each input image individually without considering relative poses between images. Looking back to the literature of Simultaneous localisation and mapping (SLAM) and visual odometry, pose estimation can significantly benefit from estimating relative poses between adjacent input frames. 
Second, the radiance field is known to suffer from \textit{shape-radiance} ambiguity~\cite{zhang2020nerf++}.
Estimating camera parameters jointly with NeRF adds another degree of ambiguity, resulting in slow convergence and unstable optimisation.

To handle the limitation of large camera motion, we seek help from monocular depth estimation~\cite{ranftl2021vision, ranftl2020towards, miangoleh2021boosting, yin2022towards}. 
Our motivation is threefold:
First, monocular depth provides strong geometry cues that are beneficial to constraint \textit{shape-radiance} ambiguity. 
Second, relative poses between adjacent depth maps can be easily injected into the training pipeline via Chamfer Distance. 
Third, monocular depth is lightweight to run and does not require camera parameters as input, in contrast to multi-view stereo depth estimation.
For simplicity, we use the term \textit{mono-depth} from now on. 

Utilising mono-depth effectively is not straightforward with the presence of scale and shift distortions. In other words, mono-depth maps are not multi-view consistent. 
Previous works ~\cite{wei2021nerfingmvs, li2021neural, gao2021dynamic} simply take mono-depth into a depth-wise loss along with NeRF training.
Instead, we propose a novel and effective way to thoroughly integrate mono-depth into our system. 
First, we explicitly optimise scale and shift parameters for each mono-depth map during NeRF training by penalising the difference between rendered depth and mono-depth. 
Since NeRF by itself is trained based on multiview consistency, this step transforms mono-depth maps to undistorted multiview consistent depth maps.
We further leverage these multiview consistent depth maps in two loss terms: 
a) a Chamfer Distance loss between two depth maps of adjacent images, which injects relative pose to our system; and
b) a depth-based surface rendering loss, which further improves relative pose estimation.

In summary, we propose a method to jointly optimise camera poses and a NeRF from a sequence of images with large camera motion. Our system is enabled by three contributions. 
\textbf{First}, we propose a novel way to integrate mono-depth into unposed-NeRF training by explicitly modelling scale and shift distortions. 
\textbf{Second}, we supply relative poses to the camera-NeRF joint optimisation via an inter-frame loss using undistorted mono-depth maps.
\textbf{Third}, we further regularise our relative pose estimation with a depth-based surface rendering loss.

As a result, our method is able to handle large camera motion, and outperforms state-of-the-art methods by a significant margin in terms of novel view synthesis quality and camera trajectory accuracy.
\section{Related Work} \label{sec:related}

\paragraph{Novel View Synthesis.} 
While early Novel View Synthesis (NVS) approaches applied interpolations between pixels~\cite{chen1993view}, later works often rendered images from 3D reconstruction ~\cite{debevec1996modeling, buehler2001unstructured}. 
In recent years, different representations of the 3D scene are used, \eg meshes~\cite{Riegler2020FVS, riegler2021stable}, Multi-Plane Images~\cite{zhou2018stereo, tucker2020single}, layered depth~\cite{tulsiani2018layer} \etc.
Among them, NeRF~\cite{mildenhall2021nerf} has become a popular scene representation for its photorealistic rendering. 

A number of techniques are proposed to improve NeRF's performance with additional  regularisation~\cite{niemeyer2022regnerf, kim2022infonerf, zhang2022ray}, depth priors~\cite{wei2021nerfingmvs, deng2022depth, roessle2022dense, Yu2022MonoSDF}, surface enhancements~\cite{oechsle2021unisurf, yariv2021volume, wang2021neus} or latent codes~\cite{wang2021ibrnet, yu2021pixelnerf, trevithick2021grf}. 
Other works~\cite{mueller2022instant, yu_and_fridovichkeil2021plenoxels, Chen2022ECCV, garbin2021fastnerf} have also accelerated NeRF training and rendering.
However, most of these approaches require pre-computed camera parameters obtained from SfM algorithms~\cite{hartley2003multiple, schonberger2016structure}.



\paragraph{NeRF With Pose Optimisation.} 
Removing camera parameter preprocessing is an active line of research recently. 
One category of the methods~\cite{Sucar:etal:ICCV2021, Zhu2022CVPR,rosinol2022nerf} use a SLAM-style pipeline, that either requires RGB-D inputs or relies on accurate camera poses generated from the SLAM tracking system. Another category of works optimises camera poses with the NeRF model directly.
We term this type of method as \emph{unposed-NeRF} in this paper.
iNeRF~\cite{yen2021inerf} shows that poses for novel view images can be estimated using a reconstructed NeRF model. 
GNeRF~\cite{meng2021gnerf} combines Generative Adversarial Networks with NeRF to estimate camera poses but requires a known sampling distribution for poses. 
More relevant to our work, NeRFmm~\cite{wang2021nerfmm} jointly optimises both camera intrinsics and extrinsics alongside NeRF training. 
BARF~\cite{lin2021barf} proposes a coarse-to-fine positional encoding strategy for camera poses and NeRF joint optimisation.
SC-NeRF~\cite{jeong2021self} further parameterises camera distortion and employs a geometric loss to regularise rays. 
GARF~\cite{chng2022garf} shows that using Gaussian-MLPs makes joint pose and scene optimisation easier and more accurate. 
Recently, SiNeRF~\cite{xia2022sinerf} uses SIREN~\cite{sitzmann2020implicit} layers and a novel sampling strategy to alleviate the sub-optimality of joint optimisation in NeRFmm. 
Although showing promising results on the forward-facing dataset like LLFF~\cite{mildenhall2019local}, these approaches face difficulties when handling challenging camera trajectories with large camera motion. 
We address this issue by closely integrating mono-depth maps with the joint optimisation of camera parameters and NeRF. 
\begin{figure*}[ht]
    \centering
    \includegraphics[width=\linewidth]{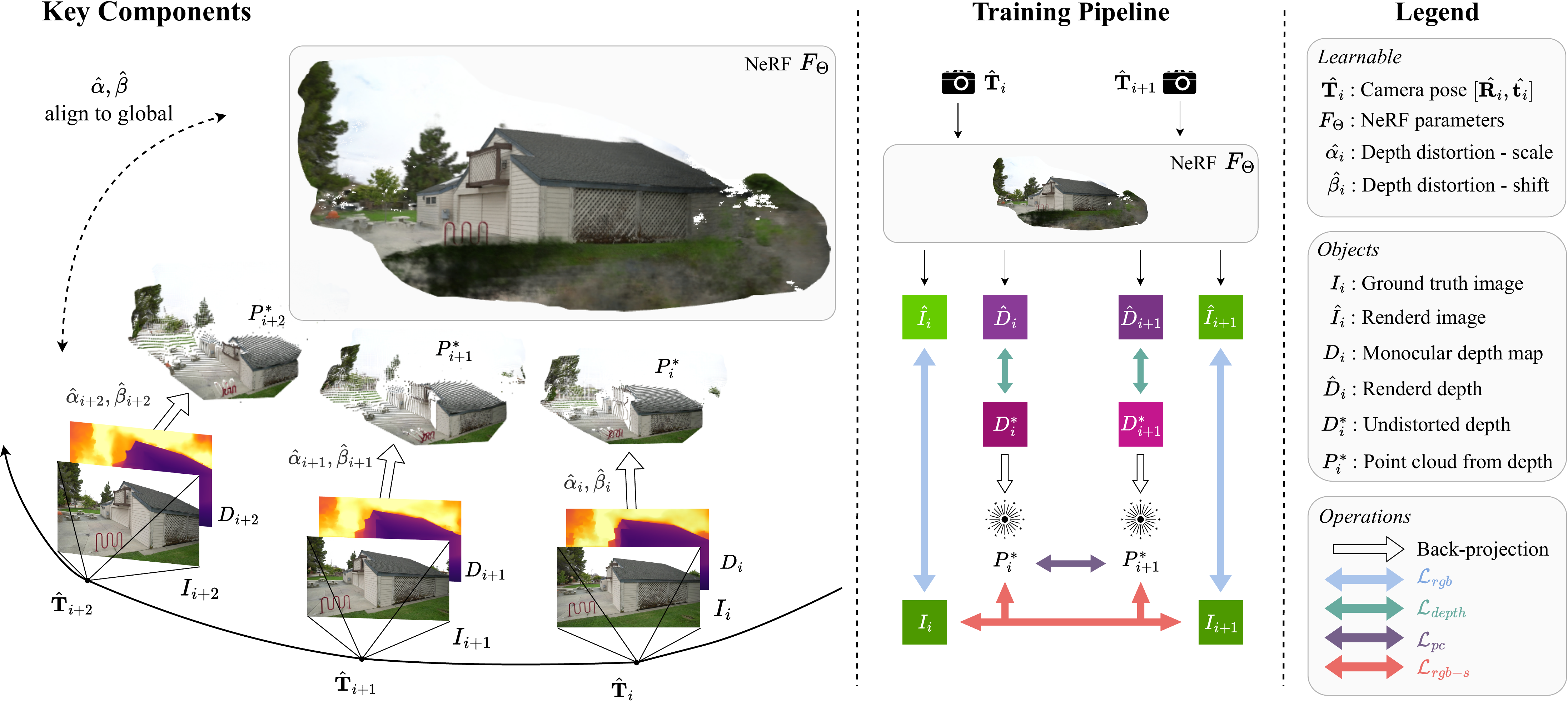}
    \caption{\textbf{Method Overview.}
    Our method takes a sequence of images as input to reconstruct NeRF and jointly estimates the camera poses of the frames. We first generate monocular depth maps from a mono-depth estimation network and reconstruct the point clouds. We then optimise NeRF, camera poses, and depth distortion parameters jointly with inter-frame and NeRF losses.
    }
    \label{fig:model}
\end{figure*}

\section{Method}\label{sec:method}
We tackle the challenge of handling large camera motions in unposed-NeRF training. 
Given a sequence of images, camera intrinsics, and their mono-depth estimations, our method recovers camera poses and optimises a NeRF simultaneously. 
We assume camera intrinsics are available in the image meta block, and we run an off-the-shelf mono-depth network DPT~\cite{deng2022depth} to acquire mono-depth estimations. 
Without repeating the benefit of mono-depth, we unroll this section around the effective integration of monocular depth into unposed-NeRF training.

The training is a joint optimisation of the NeRF, camera poses, and distortion parameters of each mono-depth map.
The distortion parameters are supervised by minimising the discrepancies between the mono-depth maps and depth maps rendered from the NeRF, which are multiview consistent.
The undistorted depth maps in return effectively mediate the \textit{shape-radiance} ambiguity, which eases the training of NeRF and camera poses.

Specifically, the undistorted depth maps enable two constraints.
We constrain global pose estimation by supplying relative pose between adjacent images. 
This is achieved via a Chamfer-Distance-based correspondence between two point clouds, back-projected from undistorted depth maps.
Further, we regularise relative pose estimation with a surface-based photometric consistency where we treat undistorted depth as surface.

We detail our method in the following sections, starting from NeRF in \cref{sec:methodnerf} and unposed-NeRF training in \cref{sec:method_nerfmm}, looking into mono-depth distortions in \cref{sec:method_undistort}, followed by our mono-depth enabled loss terms in \cref{sec:method_relative_loss}, and finishing with an overall training pipeline~\cref{sec:method_overall_pipeline}.

\subsection{NeRF}
\label{sec:methodnerf}
Neural Radiance Field (NeRF)~\cite{mildenhall2021nerf} represents a scene as a mapping function $F_\Theta: (\mathbf{x}, \mathbf{d}) \rightarrow (\mathbf{c}, \sigma)$ that maps a 3D location $\mathbf{x} \in \mathbb{R}^3$ and a viewing direction $\mathbf{d} \in \mathbb{R}^3$ to a radiance colour $\mathbf{c} \in \mathbb{R}^3$ and a volume density value $\sigma$.
This mapping is usually implemented with a neural network parameterised by $F_\Theta$.
Given $N$ images $\mathcal{I} = \left\{I_{i} \mid i=0 \ldots N-1 \right\}$ with their camera poses $\Pi = \{\pi_{i} \mid i=0 \ldots N-1\}$, NeRF can be optimised by minimising photometric error $\mathcal{L}_{rgb}=\sum_{i}^{N}\|I_{i}-\hat{I}_{i}\|_{2}^{2}$ between synthesised images $\hat{\mathcal{I}}$ and captured images $\mathcal{I}$:
    \begin{equation}\label{eq:nerf_argmin}
        \Theta^{*} =\arg \min _{\Theta} \mathcal{L}_{rgb}(\hat{\mathcal{I}} \mid \mathcal{I}, \Pi),
    \end{equation}
where $\hat{I}_{i}$ is rendered by aggregating radiance colour on camera rays $\mathbf{r}(h)=\mathbf{o}+h \mathbf{d}$ between near and far bound $h_n$ and $h_f$. 
More concretely, we synthesise $\hat{I}_{i}$ with a volumetric rendering function
    \begin{equation}\label{eq:nerf}
        \hat{I}_{i}(\mathbf{r})=\int_{h_{n}}^{h_{f}} T(h) \sigma(\mathbf{r}(h)) \mathbf{c}(\mathbf{r}(h), \mathbf{d}) d h,
    \end{equation}
where $T(h)=\exp (-\int_{h_{n}}^{h} \sigma(\mathbf{r}(s)) d s)$ is the accumulated transmittance along a ray. We refer to \cite{mildenhall2021nerf} for further details.

\subsection{Joint Optimisation of Poses and NeRF}
\label{sec:method_nerfmm}
Prior works~\cite{wang2021nerfmm,lin2021barf,jeong2021self} show that it is possible to estimate both camera parameters and a NeRF at the same time by minimising the above photometric error $\mathcal{L}_{rgb}$ under the same volumetric rendering process in~\cref{eq:nerf}. 

The key lies in conditioning camera ray casting on variable camera parameters $\Pi$, as the camera ray $\mathbf{r}$ is a function of camera pose.
Mathematically, this joint optimisation can be formulated as:
    \begin{equation}\label{eq:unpose_nerf_argmin}
        \Theta^{*}, \Pi^{*}=\arg \min _{\Theta, \Pi} \mathcal{L}_{rgb}(\hat{\mathcal{I}}, \hat{\Pi} \mid \mathcal{I}), 
    \end{equation}
where $\hat{\Pi}$ denotes camera parameters that are updated during optimising. Note that the only difference between \cref{eq:nerf_argmin} and \cref{eq:unpose_nerf_argmin} is that \cref{eq:unpose_nerf_argmin} considers camera parameters as variables.

In general, the camera parameters $\Pi$ include camera intrinsics, poses, and lens distortions.
We only consider estimating camera poses in this work, \eg, camera pose for frame $I_i$ is a transformation $\mathbf{T}_i = [\mathbf{R}_i\mid\mathbf{t}_i]$ with a rotation $\mathbf{R}_i \in \text{SO}(3)$ and a translation $\mathbf{t}_i \in \mathbb{R}^3$. 

\subsection{Undistortion of Monocular Depth}
\label{sec:method_undistort}
With an off-the-shelf monocular depth network, \eg, DPT~\cite{ranftl2021vision}, we generate mono-depth sequence $\mathcal{D}= \left\{D_{i} \mid i=0 \ldots N-1 \right\}$ from input images. 
Without surprise, mono-depth maps are not multi-view consistent so we aim to recover a sequence of multi-view consistent depth maps, which are further leveraged in our relative pose loss terms.

Specifically, we consider two linear transformation parameters for each mono-depth map, resulting in a sequence of transformation parameters for all frames $\Psi = \left\{(\alpha_{i}, \beta_{i}) \mid i=0 \ldots N-1 \right\}$, where $\alpha_{i}$ and $\beta_{i}$ denote a scale and a shift factor.
With multi-view consistent constraint from NeRF, we aim to recover a multi-view consistent depth map $D^{*}_{i}$ for $D_i$:
    \begin{equation}
    \label{eq:depth_w_distort}
        D^{*}_{i} = \alpha_{i} D_{i} + \beta_{i},
    \end{equation}
by joint optimising $\alpha_{i}$ and $\beta_{i}$ along with a NeRF. 
This joint optimisation is mostly achieved by enforcing the consistency between an undistorted depth map $D^*_i$ and a NeRF rendered depth map $\hat{D}_{i}$ via a depth loss:
    \begin{equation}
    \label{eq:loss-depth}
        \mathcal{L}_{depth} = \sum_{i}^{N}\left\|D_{i}^{*}-\hat{D}_{i}\right\|,
    \end{equation}
where
    \begin{equation}
        \hat{D}_{i}(\mathbf{r})=\int_{h_{n}}^{h_{f}} T(h) \sigma(\mathbf{r}(h)) d h
    \end{equation}
denotes a volumetric rendered depth map from NeRF.

It is important to note that both NeRF and mono-depth benefit from \cref{eq:loss-depth}. 
On the one hand, mono-depth provides strong geometry prior for NeRF training, reducing \textit{shape-radiance} ambiguity. 
On the other hand, NeRF provides multi-view consistency so we can recover a set of multi-view consistent depth maps for relative pose estimations.





\subsection{Relative Pose Constraint}
\label{sec:method_relative_loss}
Aforementioned unposed-NeRF methods \cite{wang2021nerfmm,lin2021barf,jeong2021self}
optimise each camera pose independently, resulting in an overfit to target images with incorrect poses. 
Penalising incorrect relative poses between frames can help to regularise the joint optimisation towards smooth convergence, especially in a complex camera trajectory.
Therefore, we propose two losses that constrain relative poses.

\paragraph{Point Cloud Loss.}
 We back-project the undistorted depth maps $\mathcal{D}^*$ using the known camera intrinsics,  to point clouds $\mathcal{P}^*= \left\{P^*_{i} \mid i=0 \ldots N-1 \right\}$ and optimise the relative pose between consecutive point clouds by minimising a point cloud loss $\mathcal{L}_{pc}$: 
    \begin{equation}
        \mathcal{L}_{pc} = \sum_{(i,j)}l_{cd}(P^{*}_{j}, \mathbf{T}_{ji}{P^{*}_{i}}),
    \end{equation}
where $\mathbf{T}_{ji} = \mathbf{T}_{j}\mathbf{T}_{i}^{-1}$ represents the related pose that transforms point cloud $P^{*}_i$ to $P^{*}_j$, tuple $(i, j)$ denotes indices of a consecutive pair of instances,
and $l_{cd}$ denotes Chamfer Distance:
    \begin{equation}
        \label{eq: CD}
        l_{cd}({P_{i}}, {P_{j}}) = \sum_{p_{i} \in P_{i}} \min_{p_{j} \in P_{j}} \|p_{i}-p_{j} \|_2 + \sum_{p_{j} \in P_{j}} \min_{p_{i} \in P_{i}} \|p_{i}-p_{j} \|_2.
    \end{equation}



\paragraph{Surface-based Photometric Loss.}
While the point cloud loss $\mathcal{L}_{pc}$ offers supervision in terms of 3D-3D matching, we observe that a surface-based photometric error can alleviate incorrect matching.
With the photometric consistency assumption, this photometric error penalises the differences in appearance between associated pixels.
The association is established by projecting the point cloud $P_i^*$ onto images $I_i$ and $I_j$.

The surface-based photometric loss can then be defined as:
\begin{equation}
    \mathcal{L}_{rgb-s} = \sum_{(i, j)} \| I_{i}\langle \mathbf{K}_{i}P^{*}_{i}\rangle -I_{j}\langle \mathbf{K}_{j}\mathbf{T}_{j} \mathbf{T}_{i}^{-1}{P^{*}_{i}}\rangle\|,
\end{equation}
where 
$\langle \cdot \rangle$ represents the sampling operation on the image and $\mathbf{K}_i$ denotes a projection matrix for $i_{th}$ camera.

\subsection{Overall Training Pipeline}
\label{sec:method_overall_pipeline}


Assembling all loss terms, we get the overall loss function:
    \begin{equation}
    \label{eq: 4terms}
        \mathcal{L}=\mathcal{L}_{rgb} + \lambda_{1} \mathcal{L}_{depth} + \lambda_{2} \mathcal{L}_{pc} + \lambda_{3} \mathcal{L}_{rgb-s},
    \end{equation}
where $\lambda_{1}, \lambda_{2}, \lambda_{3}$ are the weighting factors for respective loss terms.
By minimising the combined of loss $\mathcal{L}$:
    \begin{equation}
        \Theta^{*}, \Pi^{*}, \Psi^{*}=\arg \min _{\Theta, \Pi, \Psi} \mathcal{L}(\hat{\mathcal{I}},\hat{\mathcal{D}}, \hat{\Pi}, \hat{\Psi} \mid \mathcal{I}, \mathcal{D}),
    \end{equation}
our method returns the optimised NeRF parameters $\Theta$, camera poses $\Pi$, and distortion parameters $\Psi$.


\section{Experiments} \label{sec:exp} 
\begin{figure*}[ht]
    \centering
    \includegraphics[width=\linewidth]{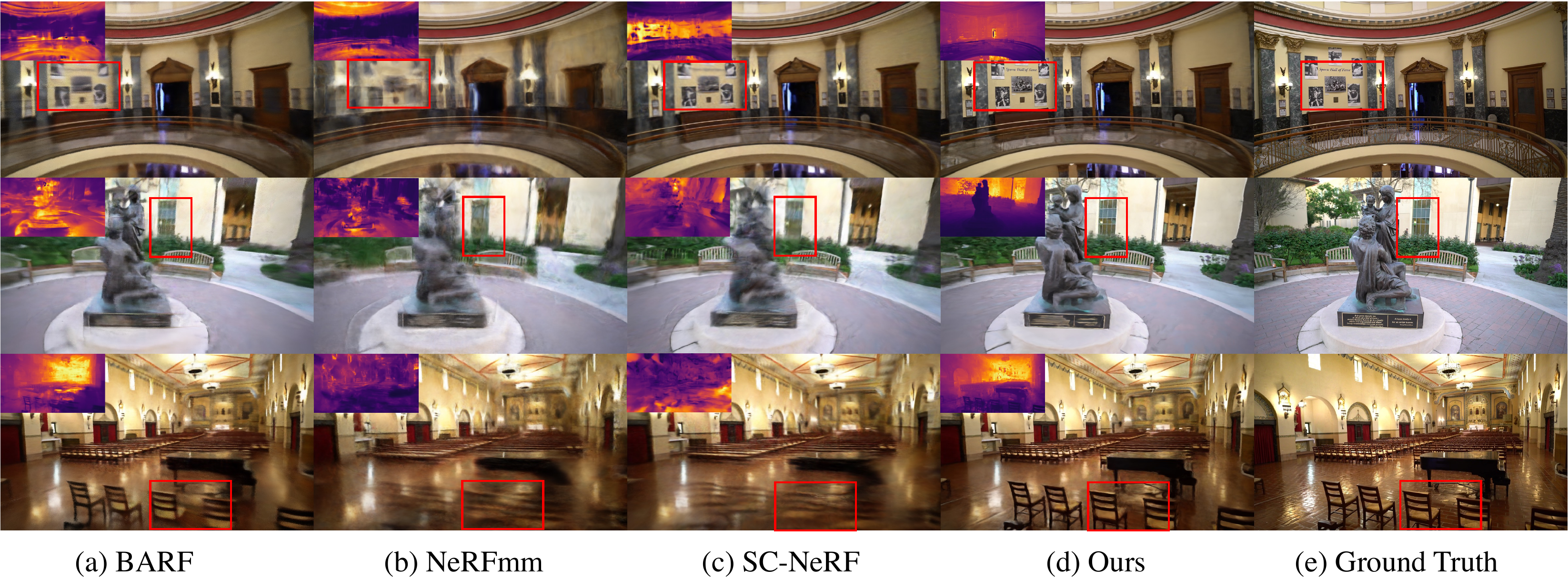}
    \caption{\textbf{Qualitative results of novel view synthesis and depth prediction on Tanks and Temples.} We visualise the synthesised images and the rendered depth maps (top left of each image) for all methods. \methodname is able to recover details for both colour and geometry.}
    \label{fig:nvs_tanks}
\end{figure*}

We begin with a description of our experimental setup in \cref{sec:exp_setup}.
In \cref{sec:acc}, we compare our method with pose-unknown methods.
Next, we compare our method with the COLMAP-assisted NeRF baseline in \cref{sec: colmap}. 
Lastly, we conduct ablation studies in \cref{sec:ablation}. 

\subsection{Experimental Setup}
\label{sec:exp_setup}
\paragraph{Datasets.} 
We conduct experiments on two datasets \textit{Tanks and Temples}~\cite{Knapitsch2017} and \textit{ScanNet}~\cite{dai2017scannet}. 
\textbf{Tanks and Temples}: we use 8 scenes to evaluate pose accuracy and novel view synthesis quality. We chose scenes captured at both indoor and outdoor  locations, with different frame sampling rates and lengths.
All images are down-sampled to a resolution of $960 \times 540$.
For the \textit{family} scene, we sample 200 images and take 100 frames with odd frame ids as training images and the remaining 100 frames for novel view synthesis, in order to analyse the performance under smooth motion.
For the remaining scenes, following NeRF~\cite{mildenhall2021nerf}, 1/8 of the images in each sequence are held out for novel view synthesis, unless otherwise specified.
\textbf{ScanNet}: we select 4 scenes for evaluating pose accuracy,
depth accuracy, and novel view synthesis quality. 
For each scene, we take 80-100 consecutive images and use 1/8 of these images for novel view synthesis. 
For evaluation, we employ depth maps and poses provided by ScanNet as ground truth.
ScanNet images are down-sampled to $648 \times 484$. We crop images with dark orders during preprocessing.

\paragraph{Metrics.}
We evaluate our proposed method in three aspects. For \textbf{novel view synthesis}, we follow previous methods~\cite{lin2021barf, wang2021nerfmm, jeong2021self}, 
and use standard evaluation metrics, including Peak Signal-to-Noise Ratio (PSNR), Structural Similarity Index Measure (SSIM)~\cite{wang2004image} and Learned Perceptual Image Patch Similarity (LPIPS)~\cite{zhang2018unreasonable}. 
For \textbf{pose} evaluation, We use standard visual odometry metrics~\cite{sturm2012benchmark,zhang2018tutorial, kopf2021robust}, including the Absolute Trajectory Error (ATE) and Relative Pose Error (RPE). ATE measures the difference between the estimated camera positions and the ground truth positions. RPE measures the relative pose errors between pairs of images, which consists of relative rotation error ($\text{RPE}_r$) and relative translation error ($\text{RPE}_t$). The estimated trajectory is aligned with the ground truth using Sim(3) with 7 degrees of freedom.
We use standard depth metrics \cite{sun2022sc, liu2015learning, fu2018deep, luo2020consistent} (Abs Rel, Sq Rel, RMSE, RMSE log, $\delta_1$, $\delta_2$ and $\delta_3$) for \textbf{depth} evaluation. For further detail, please refer to the supplementary material.
To recover the metric scale, we follow Zhou \etal \cite{zhou2017unsupervised} and match the median value between rendered and ground truth depth maps.

\paragraph{Implementation Details.}
Our model architecture is based on NeRF~\cite{mildenhall2021nerf} with a few modifications: a) replacing ReLU activation function with Softplus and b) sampling 128 points along each ray uniformly with noise, between a predefined range $(0.1, 10)$. 
We use 2 separate Adam optimisers~\cite{DBLP:journals/corr/KingmaB14} for NeRF and other parameters. 
The initial learning rate for NeRF is 0.001 and for the pose and distortion is 0.0005. 
Camera rotations are optimised in axis-angle representation $\boldsymbol{\phi}_i \in \mathfrak{so}(3)$.
We first train the model with all losses with constant learning rates until convergence.
Then, we decay the learning rates with different schedulers and gradually reduce weights for the inter-frame losses and depth loss to further train for 10,000 epochs.
We balance the loss terms with $\lambda_1=0.04$, $\lambda_2=1.0$ and $\lambda_3=1.0$.
For each training step, we randomly sample 1024 pixels (rays) from each input image and 128 samples per ray. 
More details are provided in the supplementary material.
\begin{figure*}[ht]
    \centering
    \includegraphics[width=0.9\linewidth]{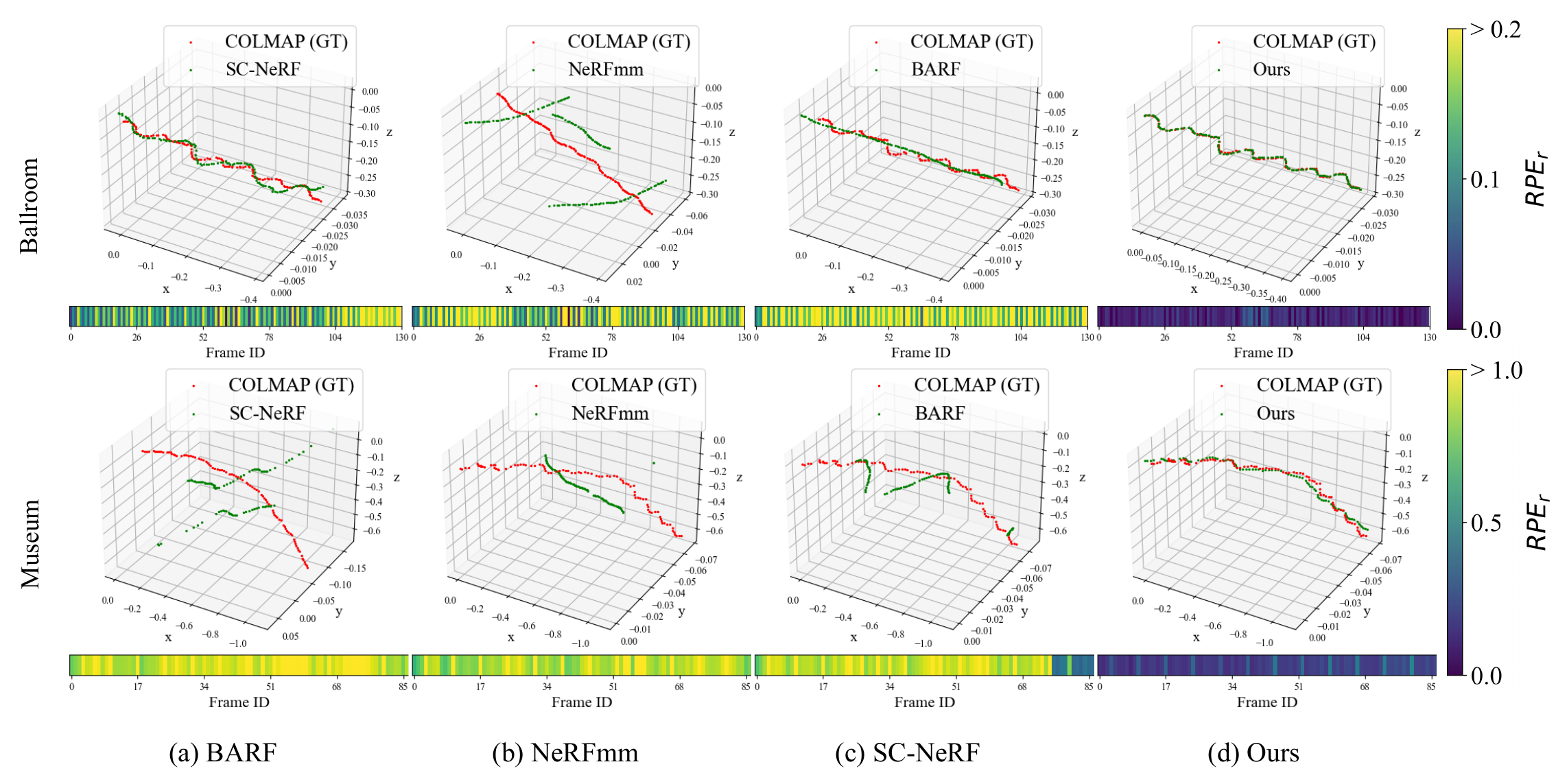}
    \caption{\textbf{Pose Estimation Comparison.} We visualise the trajectory (3D plot) and relative rotation errors $\text{RPE}_r$ (bottom colour bar) of each method on \textit{Ballroom} and \textit{Museum}. The colour bar on the right shows the relative scaling of colour. More results are in the supplementary.
    }
    \label{fig:traj1}
\end{figure*}
\subsection{Comparing With Pose-Unknown Methods}
\label{sec:acc}
We compare our method with pose-unknown baselines,
including BARF~\cite{lin2021barf}, NeRFmm~\cite{wang2021nerfmm} and SC-NeRF\cite{jeong2021self}.

\begin{table*}[ht]
\centering
\footnotesize
\begin{tabular}{cccccccccccccccccc}
\hline
                                      & \multirow{2}{*}{scenes} &  & \multicolumn{3}{c}{Ours}                       &  & \multicolumn{3}{c}{BARF}     &  & \multicolumn{3}{c}{NeRFmm} &  & \multicolumn{3}{c}{SC-NeRF} \\ \cline{4-6} \cline{8-10} \cline{12-14} \cline{16-18} 
                                      &                         &  & PSNR $\uparrow$           & SSIM $\uparrow$          & LPIPS $\downarrow$         &  & PSNR  & SSIM & LPIPS        &  & PSNR  & SSIM   & LPIPS &  & PSNR    & SSIM  & LPIPS  \\ \hline
\multicolumn{1}{c}{\multirow{5}{*}{\rotatebox[origin=c]{90}{ScanNet}}} & 0079\_00                &  & \textbf{32.47} & \textbf{0.84} & \textbf{0.41} &  & 32.31 & 0.83          & 0.43  &  & 30.59   & 0.81    & 0.49   &  & 31.33 & 0.82 & 0.46          \\
\multicolumn{1}{c}{}                  & 0418\_00                &  & \textbf{31.33} & \textbf{0.79} & \textbf{0.34} &  & 31.24 & \textbf{0.79} & 0.35  &  & 30.00   & 0.77    & 0.40   &  & 29.05 & 0.75 & 0.43          \\
\multicolumn{1}{c}{}                  & 0301\_00                &  & \textbf{29.83} & \textbf{0.77} & 0.36          &  & 29.31 & 0.76          & 0.38  &  & 27.84   & 0.72    & 0.45   &  & 29.45 & \textbf{0.77} & \textbf{0.35} \\
\multicolumn{1}{c}{}                  & 0431\_00                &  & \textbf{33.83} & \textbf{0.91} & \textbf{0.39} &  & 32.77 & 0.90          & 0.41  &  & 31.44   & 0.88    & 0.45   &  & 32.57 & 0.90 & 0.40          \\
\multicolumn{1}{c}{}                  & mean                    &  & \textbf{31.86} & \textbf{0.83} & \textbf{0.38} &  & 31.41 & 0.82          & 0.39  &  & 29.97   & 0.80    & 0.45   &  & 30.60 & 0.81 & 0.41          \\ \hline
\multicolumn{1}{c}{\multirow{9}{*}{\rotatebox[origin=c]{90}{Tanks and Temples}}}                     & Church                  &  & \textbf{25.17} & \textbf{0.73} & \textbf{0.39} &  & 23.17 & 0.62          & 0.52  &  & 21.64   & 0.58    & 0.54   &  & 21.96 & 0.60 & 0.53          \\
                                      & Barn                    &  & \textbf{26.35} & \textbf{0.69} & \textbf{0.44} &  & 25.28 & 0.64          & 0.48  &  & 23.21   & 0.61    & 0.53   &  & 23.26 & 0.62 & 0.51          \\
                                      & Museum                  &  & \textbf{26.77} & \textbf{0.76} & \textbf{0.35} &  & 23.58 & 0.61          & 0.55  &  & 22.37   & 0.61    & 0.53   &  & 24.94 & 0.69 & 0.45          \\
                                      & Family                  &  & \textbf{26.01} & \textbf{0.74} & \textbf{0.41} &  & 23.04 & 0.61          & 0.56  &  & 23.04   & 0.58    & 0.56   &  & 22.60 & 0.63 & 0.51          \\
                                      & Horse                   &  & \textbf{27.64} & \textbf{0.84} & \textbf{0.26} &  & 24.09 & 0.72          & 0.41  &  & 23.12   & 0.70    & 0.43   &  & 25.23 & 0.76 & 0.37          \\
                                      & Ballroom                &  & \textbf{25.33} & \textbf{0.72} & \textbf{0.38} &  & 20.66 & 0.50          & 0.60  &  & 20.03   & 0.48    & 0.57   &  & 22.64 & 0.61 & 0.48          \\
                                      & Francis                 &  & \textbf{29.48} & \textbf{0.80} & \textbf{0.38} &  & 25.85 & 0.69          & 0.57  &  & 25.40   & 00.69   & 0.52   &  & 26.46 & 0.73 & 0.49          \\
                                      & Ignatius                &  & \textbf{23.96} & \textbf{0.61} & \textbf{0.47} &  & 21.78 & 0.47          & 0.60  &  & 21.16   & 0.45    & 0.60   &  & 23.00 & 0.55 & 0.53          \\
                                      & mean                    &  & \textbf{26.34} & \textbf{0.74} & \textbf{0.39} &  & 23.42 & 0.61          & 0.54  &  & 22.50   & 0.59    & 0.54   &  & 23.76 & 0.65 & 0.48          \\ \hline
\end{tabular}
\caption{\textbf{Novel view synthesis results on ScanNet and Tanks and Temples}. Each baseline method is trained with its public code under the original settings and evaluated with the same evaluation protocol.}
\label{table:nvs}
\end{table*}

\begin{table*}[t]
\centering
\scriptsize
\begin{tabular}{cccccccccccccccccc} 
\hline
\multirow{2}{*}{} & \multirow{2}{*}{scenes} &  & \multicolumn{3}{c}{Ours}                                                 &  & \multicolumn{3}{c}{BARF}                   &  & \multicolumn{3}{c}{NeRFmm}              &  & \multicolumn{3}{c}{SC-NeRF}              \\ 
\cline{4-6}\cline{8-10}\cline{12-14}\cline{16-18}
                  &                         &  & $\text{RPE}_t \downarrow$ & $\text{RPE}_r \downarrow$ & ATE$ \downarrow$ &  & $\text{RPE}_t $ & $\text{RPE}_r $ & ATE &  & $\text{RPE}_t$ & $\text{RPE}_r$ & ATE   &  & $\text{RPE}_t$ & $\text{RPE}_r$ & ATE    \\ 
\hline
\multirow{5}{*}{\rotatebox[origin=c]{90}{ScanNet}} & 0079\_00                &  & \textbf{0.752}            & \textbf{0.204}            & \textbf{0.023}   &  & 1.110 &0.480 & 0.062  &  & 1.706          & 0.636          & 0.100 &  & 2.064          & 0.664          & 0.115  \\
                  & 0418\_00                &  & \textbf{0.455}            & \textbf{0.119}            & \textbf{0.015}   &  & 1.398 &0.538 & 0.020  &  & 1.402          & 0.460          & 0.013 &  & 1.528          & 0.502          & 0.016  \\
                  & 0301\_00                &  & \textbf{0.399}            & \textbf{0.123}            & \textbf{0.013}   &  & 1.316 &0.777 & 0.219  &  & 3.097          & 0.894          & 0.288 &  & 1.133          & 0.422          & 0.056  \\
                  & 0431\_00                &  & \textbf{1.625}            & \textbf{0.274}            & \textbf{0.069}   &  & 6.024 &0.754 & 0.168  &  & 6.799          & 0.624          & 0.496 &  & 4.110          & 0.499          & 0.205  \\
                  & mean                    &  & \textbf{0.808}            & \textbf{0.180}            & \textbf{0.030}   &  & 2.462 &0.637 & 0.117  &  & 3.251          & 0.654          & 0.224 &  & 2.209          & 0.522          & 0.098  \\ 
\hline
\multirow{9}{*}{\rotatebox[origin=c]{90}{Tanks and Temples}} & Church                  &  & \textbf{0.034}            & \textbf{0.008}            & \textbf{0.008}   &  & 0.114 &0.038 & 0.052
&  & 0.626          & 0.127          & 0.065 &  & 0.836          & 0.187          & 0.108  \\
                  & Barn                    &  & \textbf{0.046}            & \textbf{0.032}            & \textbf{0.004}   &  & 0.314 &0.265 & 0.050  &  & 1.629          & 0.494          & 0.159 &  & 1.317          & 0.429          & 0.157  \\
                  & Museum                  &  & \textbf{0.207}            & \textbf{0.202}            & \textbf{0.020}   &  & 3.442 &1.128 & 0.263   &  & 4.134          & 1.051          & 0.346 &  & 8.339          & 1.491          & 0.316  \\
                  & Family                  &  & \textbf{0.047}            & \textbf{0.015}            & \textbf{0.001}   &  & 1.371 &0.591 & 0.115  &  & 2.743 &0.537 & 0.120 &  & 1.171 &0.499 & 0.142  \\
                  & Horse                   &  & \textbf{0.179}            & \textbf{0.017}            & \textbf{0.003}   &  & 1.333 &0.394 & 0.014  &  & 1.349          & 0.434          & 0.018 &  & 1.366          & 0.438          & 0.019  \\
                  & Ballroom                &  & \textbf{0.041}            & \textbf{0.018}            & \textbf{0.002}   &  & 0.531 &0.228 & 0.018  &  & 0.449          & 0.177          & 0.031 &  & 0.328          & 0.146          & 0.012  \\
                  & Francis                 &  & \textbf{0.057}            & \textbf{0.009}            & \textbf{0.005}   &  & 1.321 &0.558 & 0.082  &  & 1.647          & 0.618          & 0.207 &  & 1.233          & 0.483          & 0.192  \\
                  & Ignatius                &  & \textbf{0.026}            & \textbf{0.005}            & \textbf{0.002}   &  & 0.736 &0.324 & 0.029  &  & 1.302          & 0.379          & 0.041 &  & 0.533          & 0.240          & 0.085  \\
                  & mean                    &  & \textbf{0.080}            & \textbf{0.038}            & \textbf{0.006}   &  & 1.046 &0.441 & 0.078   &  & 1.735 &0.477 & 0.123 &  & 1.890 &0.489 & 0.129   \\
\hline
\end{tabular}
\caption{\textbf{Pose accuracy on ScanNet and Tanks and Temples}. 
Note that we use COLMAP poses in Tanks and Temples as the ``ground truth". The unit of $\text{RPE}_r$ is in degrees, ATE is in the ground truth scale and $\text{RPE}_t$ is scaled by 100.
}
\label{table: pose}
\end{table*}

\textbf{View Synthesis Quality.}
\label{sec: nvs_eval}
To obtain the camera poses of test views for rendering,
 we minimise the photometric error of the synthesised images while keeping the NeRF model fixed, as in NeRFmm~\cite{wang2021nerfmm}.
Each test pose is initialised with the learned pose of the training frame that is closest to it. 
We use the same pre-processing for all baseline approaches, which results in higher accuracy than their original implementations.
More details are provided in the supplementary material. Our method outperforms all the baselines by a large margin. 
The quantitative results are summarised in~\cref{table:nvs},
and qualitative results are shown in~\cref{fig:nvs_tanks}.

 
We recognised that because the test views, which are sampled from videos, are close to the training views, good results may be obtained due to overfitting to the training images. 
Therefore, we conduct an additional qualitative evaluation on more novel views.
Specifically, we fit a bezier curve from the estimated training poses and sample interpolated poses for each method to render novel view videos.
Sampled results are shown in \cref{fig:novel},
and the rendered videos are in the supplementary material.
These results show that our method renders photo-realistic images consistently, while other methods generate visible artifacts.

\textbf{Camera Pose.}
 Our method significantly outperforms other baselines in all metrics. The quantitative pose evaluation results are shown in \cref{table: pose}.
For ScanNet, we use the camera poses provided by the dataset as ground truth.  
For Tanks and Temples, not every video comes with ground truth poses, so we use COLMAP estimations for reference. 
Our estimated trajectory is better aligned with the ground truth than other methods,
and our estimated rotation is two orders of magnitudes more accurate than others.
We visualise the camera trajectories and rotations in~\cref{fig:traj1}.

\textbf{Depth.}
We evaluate the accuracy of the rendered depth maps on ScanNet, which provides the ground-truth depths for evaluation.
Our rendered depth maps achieve superior accuracy over the previous alternatives. 
We also compare with the mono-depth maps estimated by DPT.
Our rendered depth maps, after undistortion using multiview consistency in the NeRF optimisation, outperform DPT by a large margin.
The results are summarised in~\cref{table:depth}, and sampled qualitative results are illustrated in~\cref{fig:nvs_tanks}.

\begin{table}[]
\setlength{\tabcolsep}{2pt}
\resizebox{\linewidth}{!}{
\begin{tabular}{lccccccc}
\hline
        & Abs Rel $\downarrow$ & Sq Rel $\downarrow$ & RMSE $\downarrow$ & RMSE log $\downarrow$ & $\delta_1 \uparrow$ & $\delta_2 \uparrow$ & $\delta_3 \uparrow$ \\ \hline
Ours    & \textbf{0.141}       & \textbf{0.137}      & \textbf{0.568}    & \textbf{0.176}        & \textbf{0.828}                   & \textbf{0.970}                     & \textbf{0.987}                      \\
BARF    & 0.376  &   0.684  &   0.990  &   0.401  &   0.490  &   0.751  &   0.884                     \\
NeRFmm  &0.590  &   1.721  &   1.672  &   0.587  &   0.316  &   0.560  &   0.743                        \\
SC-NeRF & 0.417  &   0.642  &   1.079  &   0.476  &   0.362  &   0.658  &   0.832 \\
DPT & 0.197  &   0.246  &   0.751  &   0.226  &   0.747  &   0.934  &   0.975
\\ \hline
\end{tabular}

}

\caption{\textbf{Depth map evaluation on ScanNet}. Our depth estimation is more accurate than baseline models BARF~\cite{lin2021barf}, NeRFmm~\cite{wang2021nerfmm} and SC-NeRF~\cite{jeong2021self}. Compared with DPT~\cite{zhou2017unsupervised}, we show our depth is more accurate after undistortion.}
\label{table:depth}
\end{table}


\begin{figure}[ht]
    \centering
    \includegraphics[width=\linewidth]{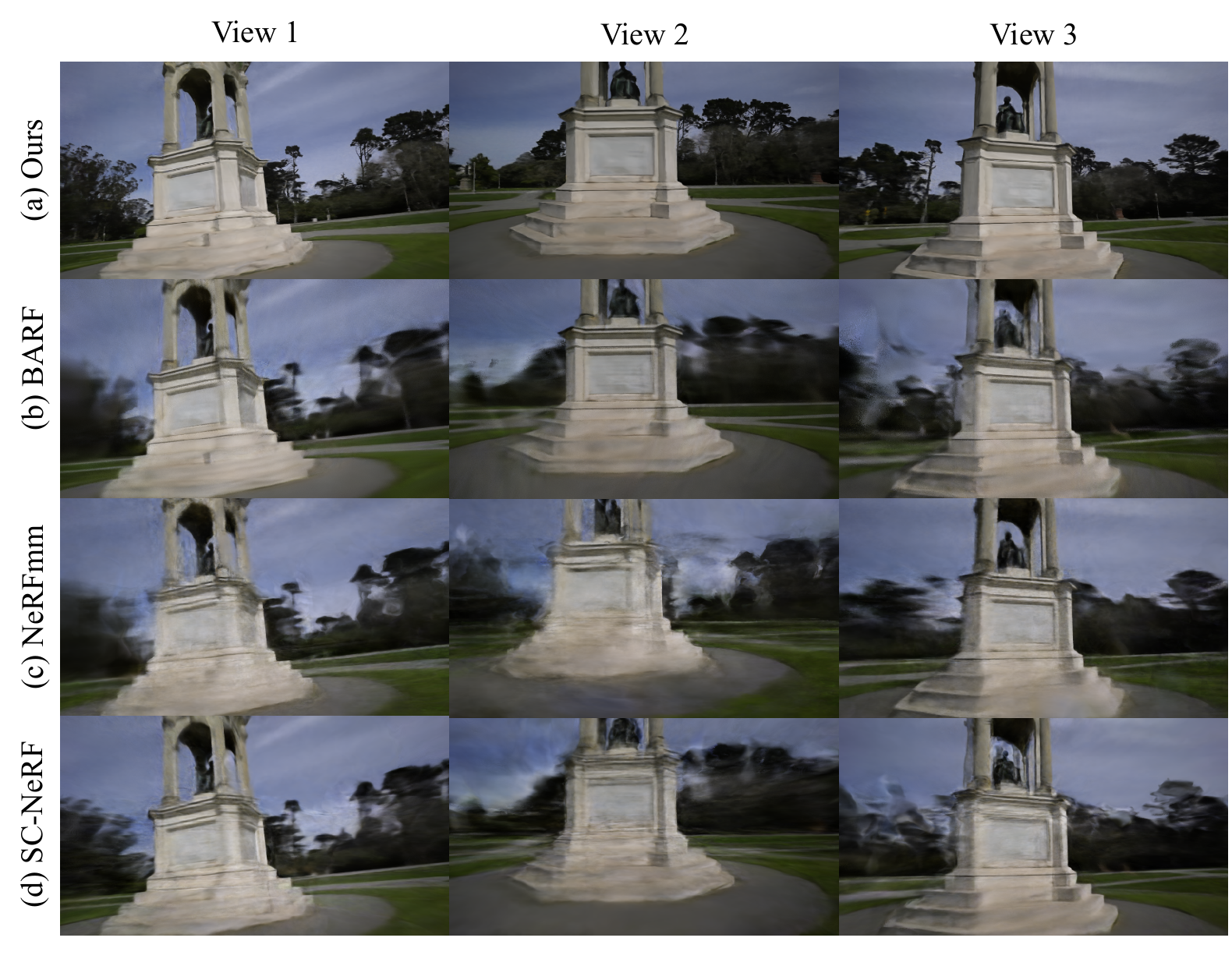}
    \caption{\textbf{Sampled frames from rendered novel view videos.} 
    For each method, we fit the learned trajectory with a bezier curve and uniformly sample new viewpoints for rendering.
    Our method generates significantly better results than previous methods, which show visible artifacts.
    The full rendered videos and details about generating novel views are provided in the supplementary.
    }
    \label{fig:novel}
\end{figure}
\subsection{Comparing With COLMAP Assisted NeRF}
\label{sec: colmap}
We make a comparison of pose estimation accuracy between our method and COLMAP against ground truth poses in ScanNet. 
We achieve on-par accuracy with COLMAP, as shown in~\cref{table: colmap pose}.  
We further analyse the novel view synthesis quality of the NeRF model trained with our learned poses to COLMAP poses on ScanNet and Tanks and Temples.
The original NeRF training contains two stages, finding poses using COLMAP and optimising the scene representation. In order to make our comparison fairer, in this section only, we mimic a similar two-stage training as the original NeRF~\cite{mildenhall2021nerf}.
In the first stage, we train our method with all losses for camera pose estimation, \ie mimicking the COLMAP processing.
Then, we fix the optimised poses and train a NeRF model from scratch,
using the same settings and loss as the original NeRF.
This evaluation enables us to compare our estimated poses to the COLMAP poses indirectly, \ie in terms of contribution to view synthesis.

Our two-stage method outperforms the COLMAP-assisted NeRF baseline,
which indicates a better pose estimation for novel view synthesis. The results are summarised in \cref{table: colmap+nerf}.

\begin{figure}[]
    \centering
    \includegraphics[width=\linewidth]{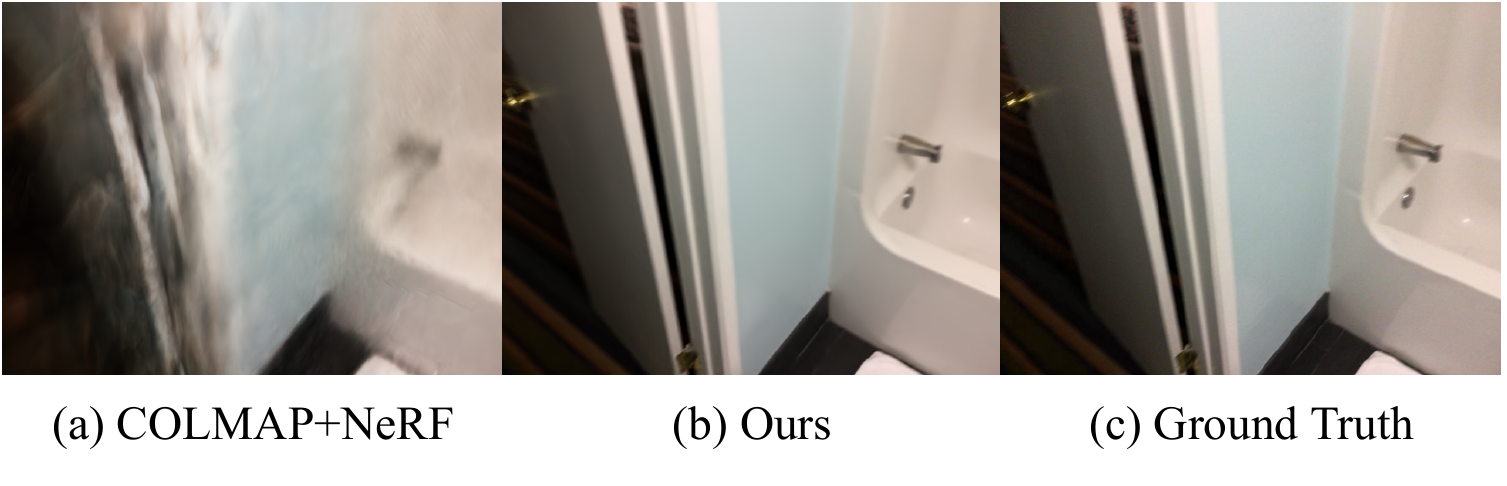}
    \caption{\textbf{COLMAP failure case.} On a rotation-dominant sequence with low-texture areas, COLMAP fails to estimate correct poses, which results in artifacts in synthesised images.}
    \label{fig:colmap_fail}
\end{figure}

As is commonly known, COLMAP performs poorly in low-texture scenes and sometimes fails to find accurate camera poses.
\cref{fig:colmap_fail} shows an example of a low-texture scene where COLMAP provides inaccurate pose estimation that causes NeRF to render images with visible artifacts.
In contrast, our method renders high-quality images, thanks to robust optimisation of camera pose.

Interestingly, this experiment also reveals that the two-stage method shows higher accuracy than the one-stage method.
We hypothesise that the joint optimisation (from randomly initialised poses) in the one-stage approach causes the NeRF optimisation to be trapped in a local minimum, potentially due to the bad pose initialisation.
The two-stage approach circumvents this issue by re-initialising the NeRF and re-training with well-optimised poses,
resulting in higher performance.

\begin{table}[]
\centering
\resizebox{\linewidth}{!}{%
\begin{tabular}{ccccccccc}
\hline
\multirow{2}{*}{scenes} &  & \multicolumn{3}{c}{Ours}                         &  & \multicolumn{3}{c}{COLMAP}                       \\ \cline{3-5} \cline{7-9} 
                        &  & $\text{RPE}_t\downarrow$        & $\text{RPE}_r\downarrow$        & ATE$\downarrow$        &  & $\text{RPE}_t$        & $\text{RPE}_r$        & ATE        \\ \hline
0079\_00                &  & 0.752          & \textbf{0.204} & 0.023          &  & \textbf{0.655} & 0.221          & \textbf{0.012} \\
0418\_00                &  & \textbf{0.455} & \textbf{0.119} & \textbf{0.015} &  & 0.491          & 0.124          & 0.016          \\
0301\_00                &  & \textbf{0.399} & \textbf{0.123} & 0.013          &  & 0.414          & 0.136          & \textbf{0.009} \\
0431\_00                &  & 1.625          & 0.274          & 0.069          &  & \textbf{1.292} & \textbf{0.249} & \textbf{0.051} \\
mean                    &  & 0.808          & \textbf{0.180} & 0.030          &  & \textbf{0.713} & 0.182          & \textbf{0.022} \\ \hline
\end{tabular}
}
\caption{\textbf{Comparison of pose accuracy with COLMAP on ScanNet.} }
\label{table: colmap pose}
\end{table}


\begin{table}[]
\centering
\setlength{\tabcolsep}{2pt}
\resizebox{\linewidth}{!}{%
\begin{tabular}{cccccccccccccc}
\hline
                                      & \multirow{2}{*}{scenes} &  & \multicolumn{3}{c}{Ours}                       &  & \multicolumn{3}{c}{Ours-r} &  & \multicolumn{3}{c}{COLMAP+NeRF}                                                      \\ \cline{4-6} \cline{8-10} \cline{12-14} 
                                      &                         &  & PSNR $\uparrow$           & SSIM $\uparrow$         & LPIPS $\downarrow$        &  & PSNR   & SSIM & LPIPS   &  & {PSNR} & {SSIM} & {LPIPS} \\ \hline
\multicolumn{1}{c}{\multirow{5}{*}{\rotatebox[origin=c]{90}{ScanNet}}} & 0079\_00                &  & 32.47          & 0.84          & 0.41          &  & \textbf{33.12} & \textbf{0.85} & \textbf{0.40} &  & 31.98                    & 0.83                     & 0.43                      \\
\multicolumn{1}{c}{}                  & 0418\_00                &  & \textbf{31.33} & \textbf{0.79} & \textbf{0.34} &  & 30.49          & 0.77          & 0.40          &  & 30.60                    & 0.78                     & 0.40                      \\
\multicolumn{1}{c}{}                  & 0301\_00                &  & 29.83          & 0.77          & 0.36          &  & \textbf{30.05} & \textbf{0.78} & \textbf{0.34} &  & 30.01                    & \textbf{0.78}            & 0.36                      \\
\multicolumn{1}{c}{}                  & 0431\_00                &  & 33.83          & 0.91          & 0.39          &  & \textbf{33.86} & \textbf{0.91} & \textbf{0.39} &  & 33.54                    & \textbf{0.91}            & \textbf{0.39}             \\
\multicolumn{1}{c}{}                  & mean                    &  & 31.86          & \textbf{0.83} & \textbf{0.38} &  & \textbf{31.88} & \textbf{0.83} & \textbf{0.38} &  & 31.53                    & 0.82                     & 0.40                      \\ \hline
\multirow{9}{*}{\rotatebox[origin=c]{90}{Tanks and Temples}}                     & Church                  &  & 25.17          & 0.73          & 0.39          &  & \textbf{26.74} & \textbf{0.78} & \textbf{0.32} &  & 25.72                    & 0.75                     & 0.37                      \\
                                      & Barn                    &  & 26.35          & 0.69          & 0.44          &  & 26.58          & \textbf{0.71} & \textbf{0.42} &  & \textbf{26.72}           & \textbf{0.71}            & \textbf{0.42}             \\
                                      & Museum                  &  & 26.77          & 0.76          & 0.35          &  & 26.98          & 0.77          & 0.36          &  & \textbf{27.21}           & \textbf{0.78}            & \textbf{0.34}             \\
                                      & Family                  &  & 26.01          & 0.74          & 0.41          &  & 26.21          & 0.75          & 0.40          &  & \textbf{26.61}           & \textbf{0.77}            & \textbf{0.39}             \\
                                      & Horse                   &  & 27.64          & \textbf{0.84} & \textbf{0.26} &  & \textbf{28.06} & \textbf{0.84} & \textbf{0.26} &  & 27.02                    & 0.82                     & 0.29                      \\
                                      & Ballroom                &  & 25.33          & 0.72          & \textbf{0.38} &  & \textbf{25.53} & \textbf{0.73} & \textbf{0.38} &  & 25.47                    & \textbf{0.73}            & \textbf{0.38}             \\
                                      & Francis                 &  & 29.48          & 0.80          & \textbf{0.38} &  & 29.73          & \textbf{0.81} & \textbf{0.38} &  & 30.05                    & \textbf{0.81}            & \textbf{0.38}             \\
                                      & Ignatius                &  & 23.96          & 0.61          & 0.47          &  & 23.98          & \textbf{0.62} & \textbf{0.46} &  & 24.08                    & 0.61                     & 0.47                      \\
                                      & mean                    &  & 26.34          & 0.74          & 0.39          &  & \textbf{26.73} & \textbf{0.75} & \textbf{0.37} &  & 26.61                    & \textbf{0.75}            & 0.38                      \\ \hline

\end{tabular}
}
\caption{\textbf{Comparison to NeRF with COLMAP poses.} Our two-stage method (Ours-r) outperforms both COLMAP+NeRF and our one-stage method (Ours).}
\label{table: colmap+nerf}
\end{table}

\subsection{Ablation Study}
\label{sec:ablation}
In this section, we analyse the effectiveness of the parameters and components that have been added to our model. The results of ablation studies are shown in \cref{table:ablation}.


\begin{table}[]
\centering
\setlength{\tabcolsep}{2pt}
\resizebox{\linewidth}{!}{%

\begin{tabular}{lccccccc} 
\hline
                      & \multicolumn{3}{c}{NVS}                               &  & \multicolumn{3}{c}{Pose}                                               \\ 
\cline{2-4}\cline{6-8}
                      & PSNR$\uparrow$ & SSIM $\uparrow$ & LPIPS $\downarrow$ &  & $\text{RPE}_t\downarrow$ & $\text{RPE}_r\downarrow$ & ATE $\downarrow$  \\ 
\hline
Ours                  & \textbf{31.86} & \textbf{0.83}   & \textbf{0.38}      &  & \textbf{0.801}           & \textbf{0.181}           & \textbf{0.031}   \\
Ours w/o $\alpha, \beta$ & 31.46          & 0.82            & 0.39               &  & 1.929                    & 0.321                    & 0.066            \\
Ours w/o $L_{pc}$       & 31.73          & 0.82            & \textbf{0.38}      &  & 2.227                    & 0.453                    & 0.101            \\
Ours w/o $L_{rgb-s}$    & 31.05          & 0.81            & 0.41               &  & 1.814                    & 0.401                    & 0.156            \\
Ours w/o $L_{depth}$    & 31.20          & 0.81            & 0.40               &  & 1.498                    & 0.383                    & 0.089            \\
\hline
\end{tabular}
}
\caption{\textbf{Ablation study results on ScanNet.}}
\label{table:ablation}
\end{table}
\textbf{Effect of Distortion Parameters.}
We find that ignoring depth distortions (\ie setting scales to 1 and shifts to 0 as constants) leads to a degradation in pose accuracy, as inconsistent distortions of depth maps introduce errors to the estimation of relative poses and confuse NeRF for geometry reconstruction.

\textbf{Effect of Inter-frame Losses.}
We observe that the inter-frame losses are the major contributor to improving relative poses. When removing the pairwise point cloud loss $L_{pc}$ or the surface-based photometric loss $L_{rgb-s}$, there is less constraint between frames, and thus the pose accuracy becomes lower. 

\textbf{Effect of NeRF Losses.}
When the depth loss $L_{depth}$ is removed, the distortions of input depth maps are only optimised locally through the inter-frame losses. We find that this can lead to drift and degradation in pose accuracy.

\subsection{Limitations}
Our proposed method optimises camera pose and the NeRF model jointly and works on challenging scenes where other baselines fail. 
However, the optimisation of the model is also affected by non-linear distortions and the accuracy of the mono-depth estimation, which we did not consider.


\section{Conclusion}
\label{sec:conclu}
In this work, we present \methodname, an end-to-end differentiable model for joint camera pose estimation and novel view synthesis from a sequence of images. We demonstrate that previous approaches have difficulty with complex trajectories. 
To tackle this challenge, we use mono-depth maps to constrain the relative poses between frames and regularise the geometry of NeRF, which leads to better pose estimation.
We show the effectiveness and robustness of \methodname on challenging scenes. The improved pose estimation leads to better novel view synthesis quality and geometry reconstruction compared with other approaches. 
We believe our method is an important step towards applying the unknown-pose NeRF models to large-scale scenes in the future. 
\section*{Acknowledgements}
We thank Theo Costain, Michael Hobley, Shuai Chen and Xinghui Li for their helpful proofreading and discussions. Wenjing Bian is supported by the China Scholarship Council - University of Oxford Scholarship.

{\small
\bibliographystyle{ieee_fullname}
\bibliography{egbib}
}
\clearpage
\begin{appendices}

\section{Implementation Details}
The following sections include more details about the datasets we use, our training procedure and evaluation metrics.

\subsection{Dataset}
We select sequences containing dramatic camera motions from ScanNet~\cite{dai2017scannet} and Tanks and Tamples~\cite{Knapitsch2017} for training and evaluation. 
\cref{table: data} lists details about these sequences, where \textit{Max rotation} denotes the maximum relative rotation angle between any two frames in a sequence. 
The sampled images are further split into training and test sets. Starting from the 5\textit{th} image, we sample every 8\textit{th} image in a sequence as a test image. 
However, this leads to a change in the sampling rate in the temporal domain among training images. 
We found that the rotation errors are often higher than average at these positions where the sampling rate changes. 
In order to study the effect of the sampling rate changes, for scene \textit{Family} in Tanks and Temples~\cite{Knapitsch2017}, we sample every other image as test images, i.e. training on images with odd frame ids and testing on images with even frame ids.

\begin{table}[ht]
\resizebox{\linewidth}{!}{%
\begin{tabular}{lccccc}
\hline
                  & Scenes   & Type    & Seq. length & Frame rate & Max. rotation (deg) \\ \hline
\multirow{4}{*}{\rotatebox[origin=c]{90}{ScanNet}} & 0079\_00 & indoor  & 90           & 30            & 54.4     \\
                  & 0418\_00 & indoor  & 80           & 30            & 27.5     \\
                  & 0301\_00 & indoor  & 100          & 30            & 43.7     \\
                  & 0431\_00 & indoor  & 100          & 30            & 45.8     \\ \hline
\multirow{8}{*}{\rotatebox[origin=c]{90}{Tanks and Temples}} & Church   & indoor  & 400          & 30            & 37.3     \\
                  & Barn     & outdoor & 150          & 10            & 47.5     \\
                  & Museum   & indoor  & 100          & 10            & 76.2     \\
                  & Family   & outdoor & 200          & 30            & 35.4     \\
                  & Horse    & outdoor & 120          & 20            & 39.0     \\
                  & Ballroom & indoor  & 150          & 20            & 30.3     \\
                  & Francis  & outdoor & 150          & 10            & 47.5     \\
                  & Ignatius & outdoor & 120          & 20            & 26.0     \\ \hline
\end{tabular}
}
\caption{\textbf{Details of selected sequences.} We downsample several videos to a lower frame rate. FPS denote frame per second. \textit{Max rotation} denotes the maximum relative rotation angle between any two frames in a sequence. We show our method can handle dramatic camera motion (large maximum rotation angle) whereas previous methods can only handle forward-facing scenes.}
\label{table: data}
\end{table}

\subsection{Training Details}
During training, we sample 1024 pixels/rays for an image and we sample 128 points along each ray for our approaches and all baselines. For all approaches, we use the same pre-defined sampling range (i.e., near and far) and sample uniformly between this range. During scheduling, the learning rate of NeRF model decays every 10 epochs with 0.9954, and the learning rate for the camera poses decays every 100 epochs with 0.9.
As the scene scales can be arbitrary, the optimised scale parameter of the depth map during training is also arbitrary. To avoid scale collapsing (all scales reduced to 0.0) during training, we manually set the scale of the depth map for the last frame to 1.0. We also use the normalised point clouds when computing the inter-frame point cloud loss.

\subsection{Test-time Optimisation}

\begin{table*}[]
\scriptsize
\centering
\begin{tabular}{lccccccccccccccc}
\hline
        & \multicolumn{3}{c}{Sim(3) + no opt.}                 &  & \multicolumn{3}{c}{Identity + opt.} &  & \multicolumn{3}{c}{Sim(3) + opt.} &  & \multicolumn{3}{c}{(4) Neighbour + opt} \\ \cline{2-4} \cline{6-8} \cline{10-12} \cline{14-16} 
        & PSNR$\uparrow$ & SSIM $\uparrow$ & LPIPS $\downarrow$ &  & PSNR          & SSIM          & LPIPS          &  & PSNR         & SSIM        & LPIPS        &  & PSNR          & SSIM          & LPIPS         \\ \hline
Ours    &  17.24              &   0.62              &   0.58                 &  &            13.38   &      0.39         &      0.70          &  &             32.47  &0.84               & 0.41             &  &             32.47  &0.84               & 0.41              \\
BARF    & 14.68                &  0.55             &    0.66              &  &            19.56   &   0.65            &  0.57              &  &              17.82&          0.60   &           0.61   &  &              32.31 &       0.83        &       0.43        \\
NeRFmm  & 11.28               &  0.40                 &  0.80                    &  &             30.59  &          0.81     &        0.49        &  &             12.46 &    0.43         &  0.80            &  &              30.59 &     0.81          &     0.49          \\
SC-NeRF & 10.68               &  0.38               &       0.80             &  &             22.39  &    0.71           &    0.55            &  &             11.25 &     0.40        &      0.80        &  &              31.33 &   0.82            &   0.46            \\ \hline
\end{tabular}
\caption{\textbf{Comparison of various pose alignment methods during test-time optimisation (ScanNet 0079\_00).}}
\label{table: test-time}
\end{table*}

During the evaluation for novel view synthesis, following our baselines NeRFmm~\cite{wang2021nerfmm}, BARF~\cite{lin2021barf} and SC-NeRF~\cite{jeong2021self}, we run a test-time optimisation to align the camera poses of the test set by minimising the photometric error on the synthesised images, while keeping the trained NeRF model froze.
Although all these baseline methods have their own way to align camera poses (discussed below), all of them fail to align camera poses in complex camera trajectories in ScanNet and Tanks and Temples. 

To fairly evaluate all methods in challenging camera trajectories, we propose to align test camera poses by first initialising from learned poses of adjacent training images, followed by a test-time optimisation. 
We shorthand this alignment as \textbf{Neighbour + opt}. 
In practice, we find this initialisation is robust and provides the best alignment for all approaches.
All results in our main paper are evaluated in this way.




The following paragraphs outline previous alignment methods, and we show a comparison for all method with a ScanNet scene in \cref{table: test-time}.

\textbf{Identity + opt.} 
BARF~\cite{lin2021barf} uses test-time optimisation to identify poses for the test frames, where all poses are initialised with identity matrices. This initialisation works well for simple forward-facing scenes, but not for complex trajectories. The optimisation is sensitive to the learning rate, and can easily fall into local minima when the target pose is far from the identity initialisation.

\textbf{Sim(3) + opt.}
In NeRFmm~\cite{wang2021nerfmm}, the poses are first initialised using Sim(3) alignment with an ATE toolbox~\cite{zhang2018tutorial}. Then, an additional test-time optimisation is used to further adjust the test poses. This initialisation works well when the learned poses can be aligned precisely to COLMAP poses (Ours in \cref{table: test-time}). However, incorrect pose estimations can affect the Sim(3) alignment.

\textbf{Sim(3) + no opt.} 
In SC-NeRF~\cite{jeong2021self}, the test poses are identified using a Sim(3) alignment between COLMAP poses and the learned poses. And no test-time optimisation is used. However, the results are biased toward COLMAP estimations, and misalignment can affect the view synthesis quality significantly.

\subsection{Evaluation Metrics}
\paragraph{Novel View Synthesis.}
We use Peak Signal-to-Noise Ratio (PSNR), Structural Similarity Index Measure (SSIM)~\cite{wang2004image} and Learned Perceptual Image Patch Similarity (LPIPS)~\cite{zhang2018unreasonable} to measure the novel view synthesis quality. For LPIPS, we use a VGG architecture~\cite{simonyan2014very}. 
\paragraph{Depth.}
The error metrics we use for depth evaluation include Abs Rel, Sq Rel, RMSE, RMSE log, $\delta_1$, $\delta_2$ and $\delta_3$. The definitions are as follows:
\begin{itemize}
    \item Abs Rel: $\frac{1}{|\mathcal{V}|} \sum_{d \in \mathcal{V}} \|d - d_{gt}\| / d_{gt}$;
    \item Sq Rel: $\frac{1}{|\mathcal{V}|} \sum_{d \in \mathcal{V}} \|d - d_{gt}\|^2_2 / d_{gt}$;
    \item RMSE: $\sqrt{\frac{1}{|\mathcal{V}|} \sum_{d \in \mathcal{V}} \|{d} -{d_{gt}}\|^2_2}$;
    \item RMSE log: $\sqrt{\frac{1}{|\mathcal{V}|} \sum_{d \in \mathcal{V}} \|\log{d} - \log{d_{gt}}\|^2_2}$;
    \item $\delta_i$: $\%$ of $y$ s.t. $ max(\frac{d}{d_{gt}}, \frac{d_{gt}}{d}) = \delta < i$;
\end{itemize}
where $d$ is the estimated depth, $d_{gt}$ is the ground truth depth, and $\mathcal{V}$ is the collection of all valid pixels on a depth map.
\section{Additional Results}
\paragraph{LLFF-NeRF Dataset.} 
We compare our approach against NeRFmm on the LLFF-NeRF dataset~\cite{mildenhall2019local} in terms of novel view synthesis quality (\cref{table: nvs_llff}) and pose accuracy (\cref{table: pose_llff}). We show better performances than NeRFmm in both pose accuracy and synthesis quality. We use the normalized device coordinate (NDC) for both approaches.
\begin{table}[]
\resizebox{\linewidth}{!}{
\begin{tabular}{cccccccccc}
\hline
                  & \multirow{2}{*}{scenes} &  & \multicolumn{3}{c}{Ours}                               &  & \multicolumn{3}{c}{NeRFmm} \\ \cline{4-6} \cline{8-10} 
                  &                         &  & PSNR $\uparrow$ & SSIM $\uparrow$ & LPIPS $\downarrow$ &  & PSNR    & SSIM   & LPIPS   \\ \hline
\multirow{9}{*}{} & Fern                    &  &                 \textbf{23.01} &\textbf{0.71}  & \textbf{0.38}      &  &       20.58 &0.59 & 0.50         \\
                  & Flower                  &  &                \textbf{29.39} &\textbf{0.86}  & \textbf{0.19}                  &  &        27.02 &0.76 & 0.32  \\
                  & Fortress                &  &                 \textbf{29.38} &\textbf{0.80}  & \textbf{0.28}                    &  &        24.94 &0.57 & 0.57   \\
                  & Horns                   &  &                 \textbf{25.24} &\textbf{0.73} & \textbf{0.37}                  &  &         23.67 &0.66  & 0.48    \\
                  & Leaves                  &  &                 \textbf{19.85} &\textbf{0.60}  & \textbf{0.40}                 &  &         19.46&0.55  & 0.46   \\
                  & Orchids                 &  &                 \textbf{19.51} &\textbf{0.56}  & \textbf{0.43}                   &  &        16.77 &0.40 & 0.55   \\
                  & Room                    &  &                 \textbf{28.54} &\textbf{0.89}  & \textbf{0.28}                  &  &         26.14 &0.84  & 0.39 \\
                  & Trex                    &  &                \textbf{25.82} &\textbf{0.84} & \textbf{0.29}                  &  &         24.13 &0.77  & 0.39 \\ \cline{2-10} 
                  & mean                    &  &              \textbf{25.09} &\textbf{0.75} & \textbf{0.33}                    &  &         22.84 &0.64 & 0.46         \\ \cline{2-10} 
\end{tabular}
}
\caption{\textbf{Novel view synthesis results on LLFF-NeRF dataset.}}
\label{table: nvs_llff}
\end{table}

\begin{table}[]
\resizebox{\linewidth}{!}{
\begin{tabular}{cccccccccc}
\hline
                  & \multirow{2}{*}{scenes} &  & \multicolumn{3}{c}{Ours}                               &  & \multicolumn{3}{c}{NeRFmm} \\ \cline{4-6} \cline{8-10} 
                  &                         &  & $\text{RPE}_t \downarrow$ & $\text{RPE}_r \downarrow$ & ATE$ \downarrow$ &  & $\text{RPE}_t $ & $\text{RPE}_r $ & ATE   \\ \hline
\multirow{9}{*}{} & Fern                    &  &   \textbf{0.252} &\textbf{0.993} & \textbf{0.003}  &  &         0.706 &1.816 & 0.007        \\
                  & Flower                  &  &                 \textbf{0.035} &\textbf{0.096} & \textbf{0.001}                   &  &         0.086 &0.418 & \textbf{0.001}    \\
                  & Fortress                &  &                \textbf{0.081} &\textbf{0.296} & \textbf{0.001}                     &  &        0.233 &0.739 & 0.004        \\
                  & Horns                   &  &                 \textbf{0.217} &\textbf{0.452} & \textbf{0.004}                   &  &         0.321 &0.850 & 0.008         \\
                  & Leaves                  &  &                 0.218 &0.143 & 0.002                   &  &         \textbf{0.138} &\textbf{0.051} & \textbf{0.001}      \\
                  & Orchids                 &  &                \textbf{0.203} &\textbf{0.383} & \textbf{0.003}                   &  &         0.686 &2.030 & 0.010         \\
                  & Room                    &  &                 \textbf{0.244} &\textbf{0.936} & \textbf{0.004}                   &  &         0.670 &1.664 & 0.011       \\
                  & Trex                    &  &                \textbf{0.219} &\textbf{0.319} & \textbf{0.004}                    &  &         0.542 &0.775 & 0.009       \\ \cline{2-10} 
                  & mean                    &  &           \textbf{0.184} &\textbf{0.452} & \textbf{0.003}                   &  &        0.423 &1.043 & 0.006        \\ \cline{2-10} 
\end{tabular}
}
\caption{\textbf{Pose accuracy on LLFF-NeRF dataset.}}
\label{table: pose_llff}
\end{table}
\paragraph{Depth Estimation.}
We show detailed depth evaluation results for ScanNet scenes in \cref{table:depth1,table:depth2,table:depth3,table:depth4}. 
Our depth estimation accuracy outperforms other baselines by a large margin. 

\begin{table}[h]
\setlength{\tabcolsep}{2pt}
\resizebox{\linewidth}{!}{
\begin{tabular}{lccccccc}
\hline
     0079\_00   & Abs Rel $\downarrow$ & Sq Rel $\downarrow$ & RMSE $\downarrow$ & RMSE log $\downarrow$ & $\delta_1 \uparrow$ & $\delta_2 \uparrow$ & $\delta_3 \uparrow$ \\ \hline
Ours    & \textbf{0.099}  &   \textbf{0.047}  &   \textbf{0.335}  &   \textbf{0.128}  &   \textbf{0.904}  &   \textbf{0.995}  &   \textbf{1.000}                      \\
BARF    & 0.208  &   0.165  &   0.588  &   0.263  &   0.639  &   0.896  &   0.983                  \\
NeRFmm  &0.494 &  1.049 &  1.419 &  0.534 &  0.378 &  0.567 &  0.765                      \\
SC-NeRF & 0.360  &   0.450  &   0.902  &   0.396  &   0.407  &   0.730  &   0.908 \\
DPT & 0.149  &   0.095  &   0.456  &   0.173  &   0.818  &   0.978  &   0.999
\\ \hline
\end{tabular}
}
\caption{\textbf{Depth map evaluation on ScanNet 0079\_00}.}
\label{table:depth1}
\end{table}

\begin{table}[h]
\setlength{\tabcolsep}{2pt}
\resizebox{\linewidth}{!}{
\begin{tabular}{lccccccc}
\hline
     0418\_00   & Abs Rel $\downarrow$ & Sq Rel $\downarrow$ & RMSE $\downarrow$ & RMSE log $\downarrow$ & $\delta_1 \uparrow$ & $\delta_2 \uparrow$ & $\delta_3 \uparrow$ \\ \hline
Ours    & \textbf{0.152}  &   \textbf{0.137}  &   \textbf{0.645}  &   \textbf{0.185}  &   \textbf{0.738}  &   \textbf{0.988}  &   \textbf{0.997}                     \\
BARF    & 0.718  &   1.715  &   1.563  &   0.630  &   0.205  &   0.569  &   0.769                    \\
NeRFmm  &0.907  &   3.650  &   2.176  &   0.769  &   0.240  &   0.456  &   0.621                       \\
SC-NeRF & 0.319  &   0.441  &   0.898  &   0.377  &   0.456  &   0.792  &   0.930 \\
DPT & 0.190  &   0.187  &   0.745  &   0.211  &   0.719  &   0.965  &   \textbf{0.997}
\\ \hline
\end{tabular}
}
\caption{\textbf{Depth map evaluation on ScanNet 0418\_00}.}
\label{table:depth2}
\end{table}

\begin{table}[h]
\setlength{\tabcolsep}{2pt}
\resizebox{\linewidth}{!}{
\begin{tabular}{lccccccc}
\hline
     0301\_00   & Abs Rel $\downarrow$ & Sq Rel $\downarrow$ & RMSE $\downarrow$ & RMSE log $\downarrow$ & $\delta_1 \uparrow$ & $\delta_2 \uparrow$ & $\delta_3 \uparrow$ \\ \hline
Ours    &  0.185  &   0.252  &   0.711  &   \textbf{0.233}  &   \textbf{0.792}  &   \textbf{0.918}  &   \textbf{0.958}                      \\
BARF    & \textbf{0.179}  &   \textbf{0.146}  &   \textbf{0.502}  &   0.268  &   0.736  &   0.883  &   0.938                   \\
NeRFmm  &0.444  &   0.830  &   1.239  &   0.481  &   0.397  &   0.680  &   0.845                     \\
SC-NeRF & 0.383  &   0.378  &   0.810  &   0.452  &   0.360  &   0.663  &   0.846 \\
DPT & 0.317  &   0.568  &   1.133  &   0.350  &   0.597  &   0.821  &   0.914
\\ \hline
\end{tabular}
}
\caption{\textbf{Depth map evaluation on ScanNet 0301\_00}.}
\label{table:depth3}
\end{table}

\begin{table}[h]
\setlength{\tabcolsep}{2pt}
\resizebox{\linewidth}{!}{
\begin{tabular}{lccccccc}
\hline
     0431\_00   & Abs Rel $\downarrow$ & Sq Rel $\downarrow$ & RMSE $\downarrow$ & RMSE log $\downarrow$ & $\delta_1 \uparrow$ & $\delta_2 \uparrow$ & $\delta_3 \uparrow$ \\ \hline
Ours    & \textbf{0.127}  &   \textbf{0.111}  &   \textbf{0.579}  &   \textbf{0.160}  &   \textbf{0.877}  &   \textbf{0.978}  &   \textbf{0.994}                     \\
BARF    & 0.398  &   0.710  &   1.307  &   0.444  &   0.381  &   0.655  &   0.847                  \\
NeRFmm  &0.514  &   1.354  &   1.855  &   0.562  &   0.250  &   0.539  &   0.742                        \\
SC-NeRF & 0.608  &   1.300  &   1.706  &   0.677  &   0.225  &   0.446  &   0.645  \\
DPT & 0.132  &   0.135  &   0.670  &   0.171  &   0.855  &   0.973  &   0.991
\\ \hline
\end{tabular}
}
\caption{\textbf{Depth map evaluation on ScanNet 0431\_00}.}
\label{table:depth4}
\end{table}

\begin{figure*}[]
    \centering
    \includegraphics[width=0.9\linewidth]{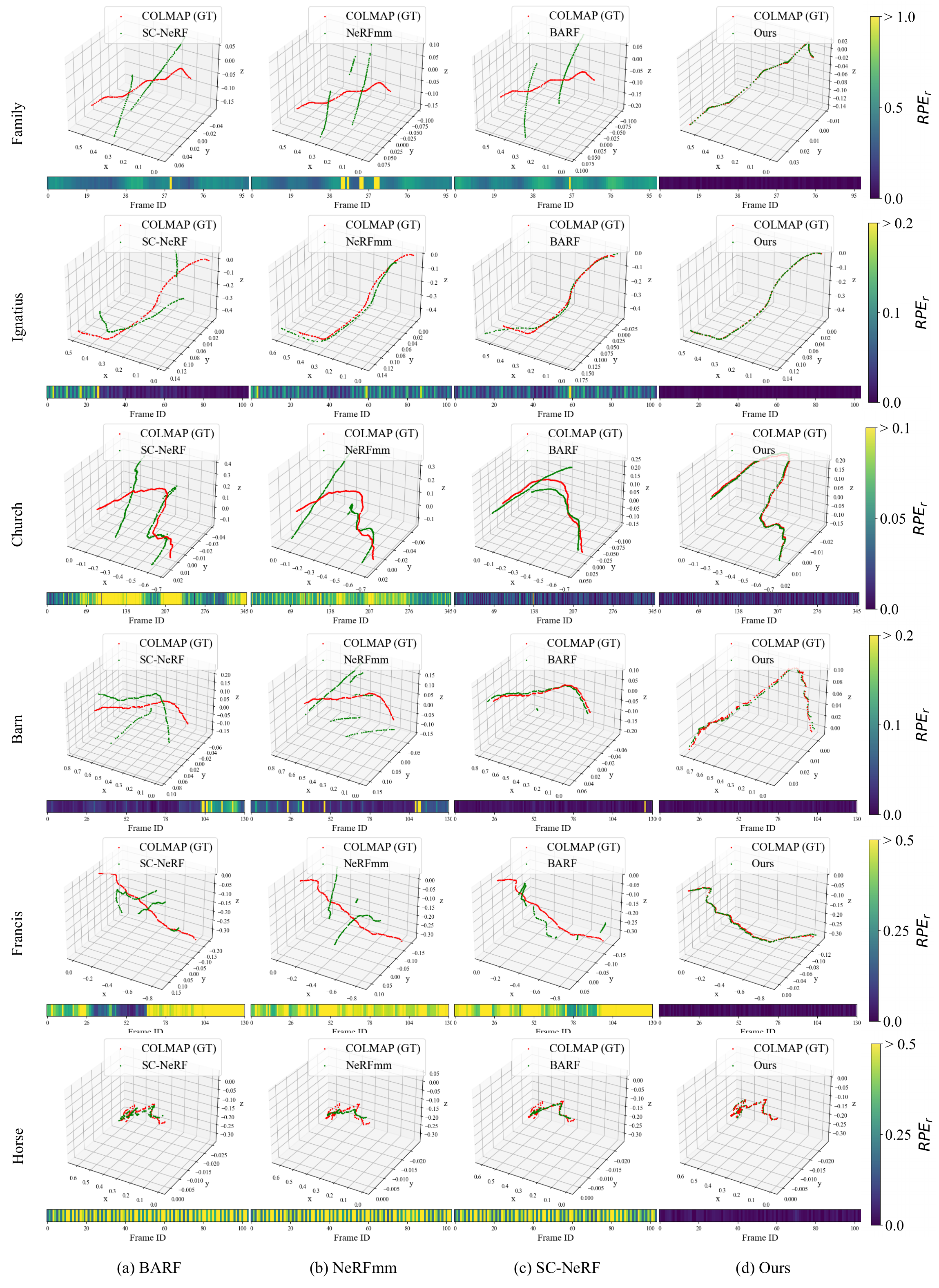}
    \caption{\textbf{Pose Estimation Comparison on Tanks and Temples.} We visualise the trajectory (3D plot) and relative rotation errors $\text{RPE}_r$ (bottom colour bar) of each method on \textit{Ballroom} and \textit{Museum}. The colour bar on the right shows the relative scaling of colour.
    }
    \label{fig:traj2}
\end{figure*}

\begin{figure*}[]
    \centering
    \includegraphics[width=0.9\linewidth]{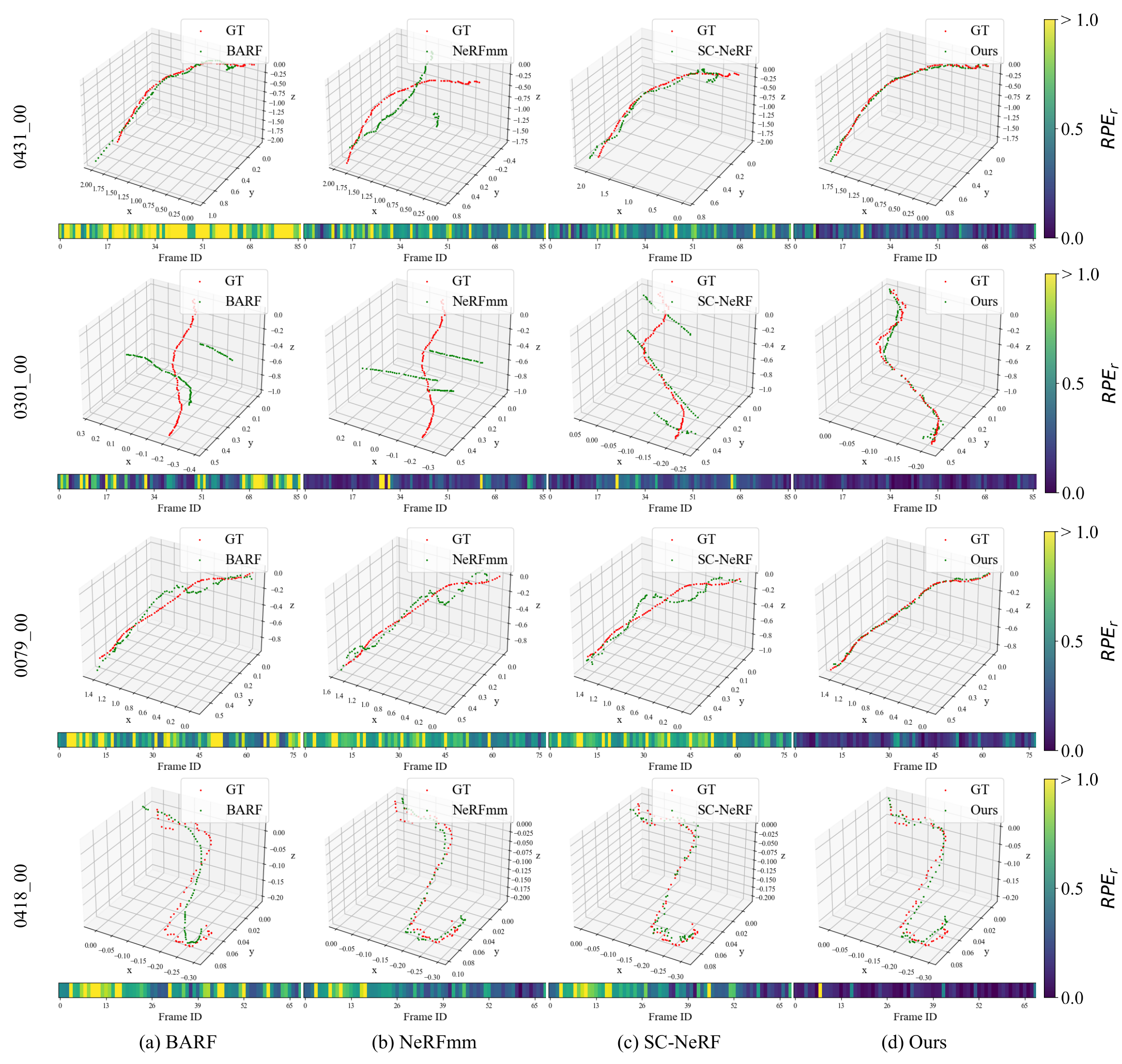}
    \caption{\textbf{Pose Estimation Comparison on ScanNet.} We visualise the trajectory (3D plot) and relative rotation errors $\text{RPE}_r$ (bottom colour bar) of each method on \textit{Ballroom} and \textit{Museum}. The colour bar on the right shows the relative scaling of colour.
    }
    \label{fig:traj_scannet}
\end{figure*}

\paragraph{Pose Estimation.}
We visualise additional results for pose estimation on Tanks and Temples (\cref{fig:traj2}) and ScanNet (\cref{fig:traj_scannet}).

\begin{figure*}[h]
    \centering    \includegraphics[width=\linewidth]{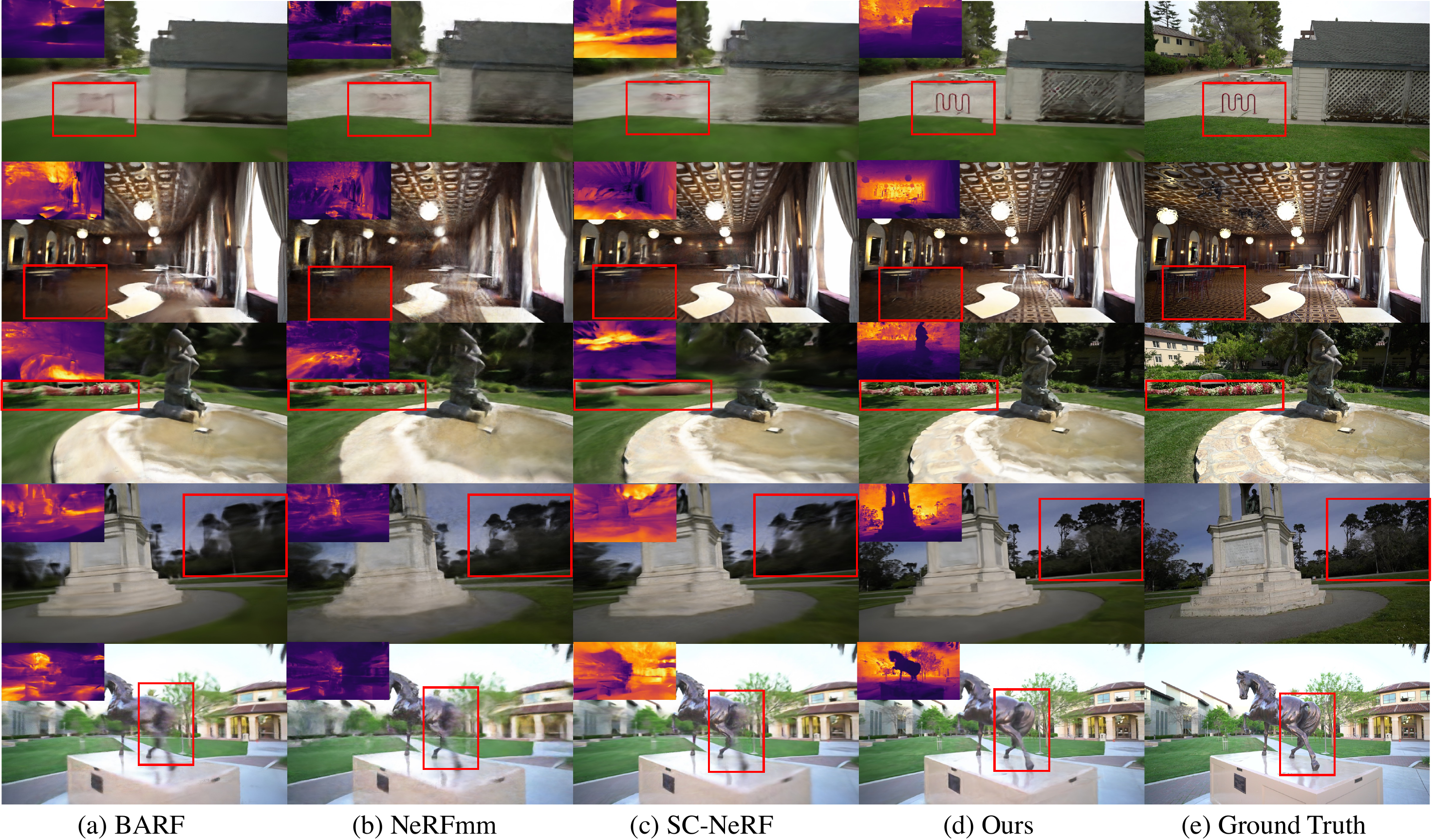}
    \caption{\textbf{Qualitative results of novel view synthesis and depth prediction on Tanks and Temples.} We visualise the synthesised images and the rendered depth maps (top left of each image) for all methods. \methodname is able to recover details for both colour and geometry.}
    \label{fig:nvs2}
\end{figure*}

\begin{figure*}[]
    \centering
    \includegraphics[width=\linewidth]{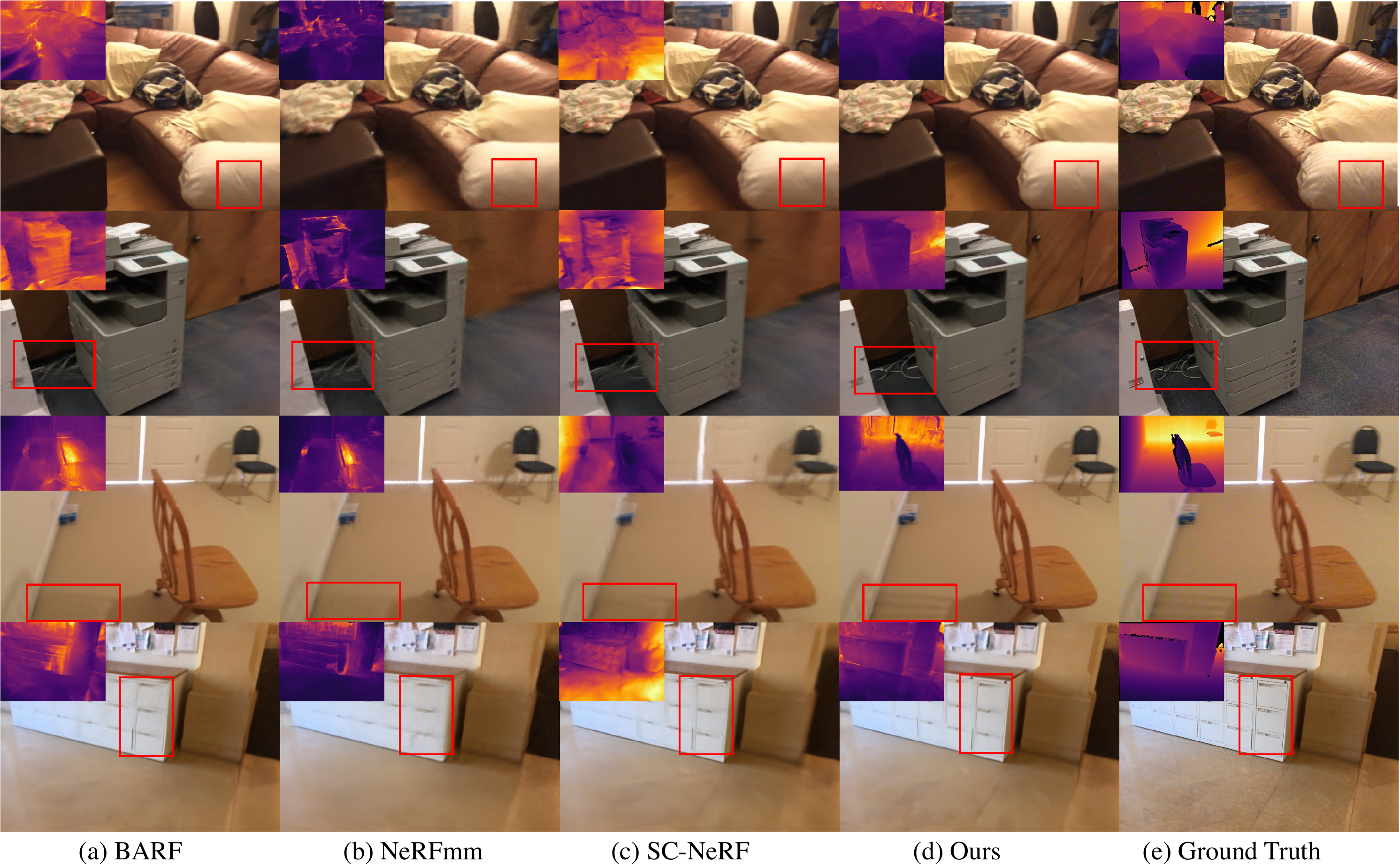}
    \caption{\textbf{Qualitative results of novel view synthesis and depth prediction on ScanNet.} We visualise the synthesised images and the rendered depth maps (top left of each image) for all methods. \methodname is able to recover details for both colour and geometry.}
    \label{fig:scannet_nvs}
\end{figure*}

\paragraph{More Visualisations.}
We present additional qualitative results for novel view synthesis and depth estimation on Tanks and Temples (\cref{fig:nvs2}) and ScanNet (\cref{fig:scannet_nvs}).

\end{appendices}
\clearpage

\end{document}